\documentclass[pdflatex,sn-mathphys-num]{sn-jnl}

\usepackage{graphicx}%
\usepackage{multirow}%
\usepackage{amsmath,amssymb,amsfonts}%
\usepackage{amsthm}%
\usepackage{mathrsfs}%
\usepackage[title]{appendix}%
\usepackage{xcolor}%
\usepackage{textcomp}%
\usepackage{manyfoot}%
\usepackage{booktabs}%
\usepackage{algorithm}%
\usepackage{algorithmicx}%
\usepackage{algpseudocode}%
\usepackage{listings}%
\usepackage{todonotes}
\usepackage{ulem}

\usepackage[T1]{fontenc}%
\usepackage{siunitx}%
\usepackage{multirow}%
\usepackage{array}%
\usepackage{tabularx}
\usepackage{placeins}
\usepackage{threeparttable}

\usepackage{subcaption}

\begin{document}

\title[Simulating temporomandibular \dots]{Simulation of temporomandibular joint space changes based on rigid-body transformation using virtual and semi-virtual models: a proof-of-concept study}

\author[1]{\fnm{Agnieszka Anna} \sur{Tomaka}}\email{ines@iitis.pl}
\author[1]{\fnm{Krzysztof} \sur{Domino}}\email{kdomino@iitis.pl}
\author[1,2]{\fnm{Michał} \sur{Tarnawski}}\email{michal@tarnawski.eu}
\author[3]{\fnm{Emil} \sur{Kalinowski}} \email{emil.kalinowski@wum.edu.pl} 
\author*[4]{\fnm{Małgorzata} \sur{Domino}}\email{malgorzata\_domino@sggw.edu.pl} 
\author*[1]{\fnm{Dariusz} \sur{Pojda}}\email{dpojda@iitis.pl}

\affil[1]{%
    \orgdiv{Institute of Theoretical and Applied Informatics},%
    \orgname{Polish Academy of Sciences},%
    \orgaddress{
        \street{Bałtycka 5},
        \postcode{44--100},
        \city{Gliwice},
        \country{Poland}
    }%
}

\affil[2]{%
    \orgdiv{Private Orthodontic Practice},%
    \orgname{Family Dentistry s.c. Barbara Michał Tarnawscy},%
    \orgaddress{
        \street{Kościuszki 44/3},
        \postcode{44-048},
        \city{Katowice},
        \country{Poland}
    }%
}

\affil[3]{%
    \orgdiv{Department of Cranio-Maxillofacial Surgery, Oral Surgery and Implantology},%
    \orgname{Medical University of Warsaw},%
    \orgaddress{
        \street{Lindleya 4},
        \postcode{02-005},
        \city{Warsaw},
        \country{Poland}
    }%
}

\affil[4]{%
    \orgdiv{Department of Large Animal Diseases and Clinic, Institute of Veterinary Medicine},%
    \orgname{Warsaw University of Life Sciences},%
    \orgaddress{
        \street{Nowoursynowska 166},
        \postcode{02-787},
        \city{Warsaw},
        \country{Poland}
    }%
}
   
\abstract{The integration of multimodal data within a common coordinate system is essential for spatial analysis of the stomatognathic system. This proof-of-concept study aims to use a multimodal data integration methodology for digital modeling of spatial changes in the temporomandibular joint (TMJ) across various mandible positions. As two digital models were investigated, the study also aimed to investigate inter-model differences in the simulated TMJ joint spaces. A transformation tree framework was proposed to integrate cone-beam computed tomography (CBCT), facial, and dental representation images into a common coordinate system and to simulate seven mandible positions using rigid-body transformations. Data were collected from a single participant. Spatial changes in mandible positions were propagated to TMJ articular surface positions using both models, although the results were model-related. The model-based uncertainty was quantified using two metrics - the consistency error and the model disagreement. The consistency error presented anisotropy in the latero-lateral direction. The model disagreement were not uniform across mandible positions and were higher (p<0.0156) at the TMJ level (1.46 mm) than at the dental level (0.47 mm). The proposed methodology may prove useful in future research investigating the geometric relationships between mandible position and the TMJ. However, given the proof-of-concept nature of this study, the numerical findings should not be generalized to broader clinical populations.}


\maketitle

\section{Introduction}

Combining data from different imaging modalities can improve the analysis of the stomatognathic system \cite{TomakaG2019}. 
Radiographic techniques provide detailed information on anatomical structures; among them, cone-beam computed tomography (CBCT) enables three-dimensional (3D) representations of maxillofacial anatomical structures, including bony structures and surrounding soft tissues
\cite{kaasalainen2021dental}. Facial morphology can be represented using two-dimensional (2D) images or 3D surface models obtained using facial scanning technique such as stereophotogrammetry \cite{heike20103d,shuto2025facial}. Detailed tooth and gingival morphology can be represented using digital dental models obtained from intraoral scans \cite{alassiry2023clinical} or optical scans of dental plaster models \cite{keating2008comparison}. 
Each imaging modality provides complementary information on different components of patient-specific anatomical structures, including teeth \cite{alassiry2023clinical, kaasalainen2021dental} and temporomandibular joints (TMJs) \cite{almashraqi2024recommendations}. Therefore, the integration of multimodal data within a common spatial coordinate system holds the potential to provide a meaningful analysis of the entire stomatognathic system~\cite{mangano2018combining,park2022creation}.

The presented study addresses the need to develop a multimodal data integration methodology to simulate mandible position and consequentially changes in spatial articular surface positions in the TMJ. The study was designed based on previous studies by our research team on multimodal data acquisition from the stomatognathic system~\cite{TomakaG2019}, mandibular motion acquisition using a facial scanner~\cite{ITIB2016, TomakaG2019, TOMAKA2025110311}, and occlusal splint design based on transformation models~\cite{Pojda2019, TOMAKA2025104527}. The proposed method represents mandibular motion as a sequence of the rigid-body transformations and uses target transformations derived from recorded mandibular motion to design occlusal splints. 
The method was subsequently expanded by applying a transformation tree framework~\cite{POJDA2025102093,TOMAKA2025110311}, which enables digital modeling of representations of selected anatomical structures, with images derived from various modalities, e.g., a CBCT scanner, a facial scanner, and an optical scanner.

Transformation trees represent hierarchical relationships between coordinate systems and store successive rigid-body transformations. Each transformation combines rotation and translation without loss of information to describe the distance and orientation of an object in 3D space while retaining its original coordinates. Such representations support the positioning and alignment of anatomical structures in medical imaging applications~\cite{Hallpike/img.14.6.140455, POJDA2025102093}.
Accordingly, mandibular motions can be represented and compared using sets of rigid-body transformations that describe the mandible in different positions. 
These transformations can then be statistically analyzed within the Lie group \(SE(3)\). In this study, repeated transformations were summarized using an iterative log-residual mean on \(SE(3)\), described below. Thus, transformation trees facilitate the integration of multimodal data across coordinate systems, enabling mandibular motion mapping, comparison of mandible positions, quantification of positioning errors~\cite{POJDA2025102093,TOMAKA2025110311}, and potentially comparison of articular surface positions.

On the one hand, previous studies have investigated the digital design of occlusal splints and the precision of mandible positioning achieved using these splints, reporting no propagation of positioning effects to the TMJ level \cite{Pojda2019,TOMAKA2025104527,wang2022coverage,vanderhelm2023validity}. On the other hand, previous researches have examined the dimensions of TMJ joint space using CBCT-based measurements \cite{IKEDA2009,Ikeda2011,henriques2012condylar,schnabl2018cbct,nota2021workflow,bk2025quantitative}, 3D surface and volumetric analyzes \cite{sagl2019disc,SHU2022107149,mackie2022quantitative,holte2023holistic,al2024comprehensive,al2025interpretable,yang2025visualization}, and mandibular motion modeling \cite{Enciso2003V,Yuuda2007,YANG2013}.
However, these two research directions have rarely been combined to assess the geometric consequences of discrepancies between planned mandible positions (digital realization) and achieved positions (physical realization using occlusal splints and plaster models) at both dental and TMJ levels. The novelty of the present study lies in integrating these two research directions within a unified computational methodology based on rigid-body transformations within a transformation tree framework. Seven predefined mandible positions are reproduced virtually (representing digital realizations) and semi-virtually (representing physical realizations). The resulting rigid-body transformations are applied to the same mandible geometry, and the discrepancies between them are evaluated at both the dental level, using surface distance maps, and the TMJ level, using surface distance maps combined with changes in TMJ joint space dimensions. This approach enables the assessment of how discrepancies in the same transformations, particularly their translation and rotation components, produce different geometric effects in these two anatomical regions.

Therefore, this study aims to apply a multimodal data integration methodology for digital modeling - both virtual and semi-virtual - of changes in the TMJ joint space across various mandible positions. It also investigates how uncertainty in mandible positioning, introduced in semi-virtual models, affects TMJ joint space simulations compared with those based on virtual models.

\section{Method}

\subsection{Study design}
This study represents a retrospective methodological research on a unified computational approach combining 
(1) multimodal data integration, 
(2) rigid-body transformations of mandible positions and related changes in TMJ joint space within the common coordinate systems, and 
(3) comparison of the transformations achieved between virtual and semi-virtual models.

\begin{figure}[ht]
\centering
    \includegraphics[width=\linewidth]{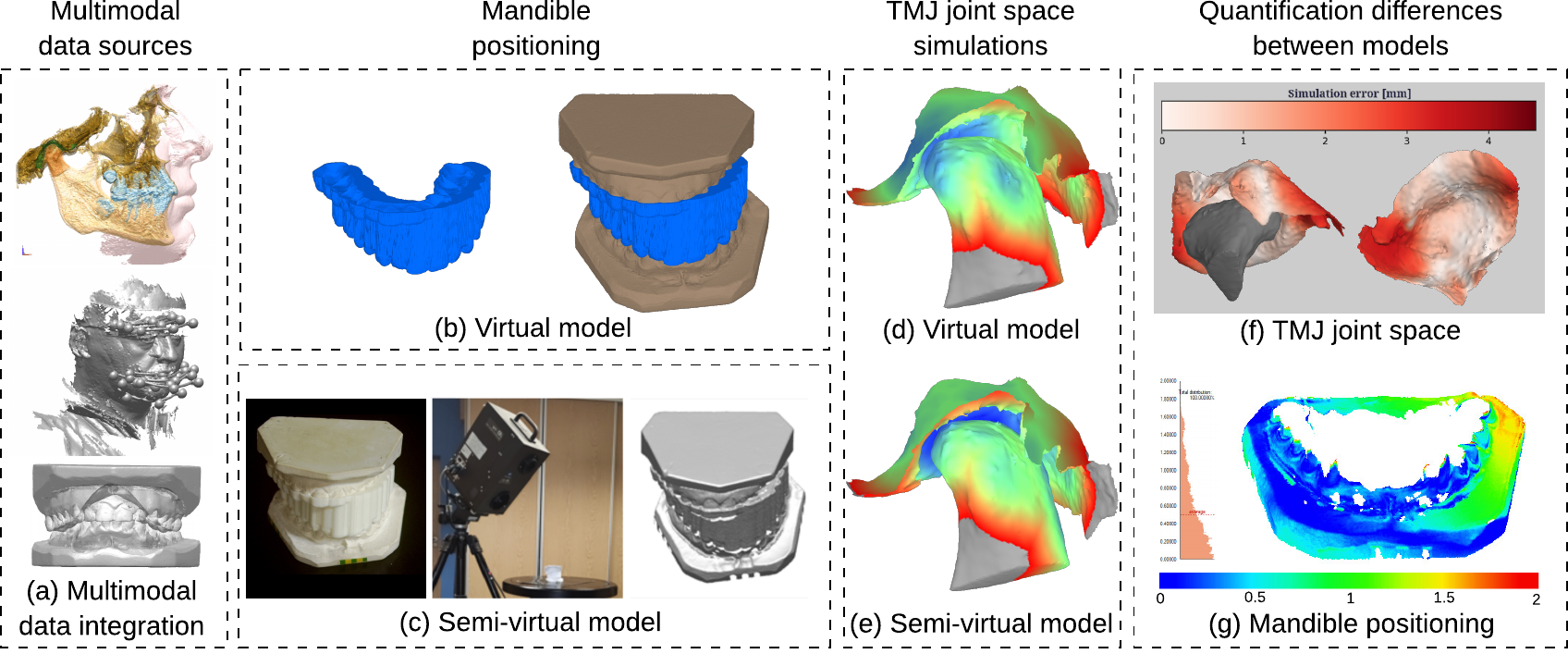}
    \caption{Schematic illustration of a study design. (a) Integration of the multimodal data collected using the cone-beam computed tomography (CBCT) scanner, facial scanner, and optical scanner. Mandible positioning using (b) the virtual model and (c) the semi-virtual model. Simulation and quantification of the temporomandibular joint (TMJ) space changes (joint space distance) using (d) the virtual model and (e) the semi-virtual model. Quantification differences between the virtual and semi-virtual models (f) at the level of the TMJs (model disagreement) and (g) at the level of mandible positioning (consistency error).}
    \label{fig:study_design}
\end{figure}

This approach enables simulations of changes in the TMJ joint space using virtual and semi-virtual models and investigates model disagreement. The study was designed in five steps, as illustrated in Figure~\ref{fig:study_design}. Digital models of changes in the TMJ joint space across different mandible positions were developed using archival CBCT images, facial images, and images of a dental plaster model (dental images) obtained from a single participant. The CBCT images, facial images, and dental plaster model were provided by the Private Orthodontic Practice (Family Dentistry s.c. Barbara Michał Tarnawscy, Katowice, Poland). Dental images were acquired at the Institute of Theoretical and Applied Informatics, Polish Academy of Sciences (Gliwice, Poland).

In the first step, CBCT and facial images were processed as 3D representations of participant-specific maxillofacial anatomical structures. Then, a dental plaster model made for this participant in an orthodontic laboratory was imaged using optical scanner to obtain 3D representation of participant-specific tooth and gingival morphology. Subsequently, three multimodal image types were integrated using the transformation tree based on the rigid-body transformation.

In the second step, facial images representing mandibular motions were recorded using the facial scanner following the previously described protocol \cite{TomakaG2019, TOMAKA2025110311}. Eight mandible positions were selected and used in further modeling. Position 0 represented the neutral (reference) position with the mouth closed and the teeth in habitual occlusion; while positions 1--7 represented the target positions with mild and moderate mouth opening with or without left or right lateral deviation or mandible protrusion, respectively. The selected mandible positions are characterized in section \ref{sec:positioning}

In the third step, the articular surfaces and mandible positions were modeled using a virtual model representing the target mandible positions. In this model, each of the target positions (positions 1--7) was defined as the rigid-body transformation relative to the reference position (position 0). Within this model, the target positions were referred to as planned positions to distinguish them from the achieved positions in the semi-virtual model. Transformation trees were used to transfer the distances and orientations of anatomical structures between the CBCT, facial, and dental coordinate systems, thereby translating changes in the planned mandible position into corresponding changes within the TMJ joint space. This approach enabled the virtual simulation of changes in the TMJ joint space within a common coordinate system.

In the fourth step, the target mandible positions were modeled using a semi-virtual model.
For this purpose, seven occlusal splints, replicating the distances and orientations of the teeth in the mandible positions 1--7, were designed and fabricated following the previously described protocol \cite{Pojda2019, TOMAKA2025104527} using dpVision software \cite{POJDA2025102093,dpvision_soft} and SplintMaker \cite{splintmaker_soft} plugin.
The occlusal splints were designed based on the planned mandible positions used solely in a virtual model. 
Each occlusal splint was then placed individually on a dental plaster model to create a physical representation of the corresponding planned mandible position. 
Each physical realization was imaged using the optical scanner following the previously described protocol \cite{TOMAKA2025110311}. 
This process yielded seven achieved mandible positions. 
The resulting digital representation of the achieved positions were transferred between the CBCT, facial, and dental coordinate systems using the transformation trees, thereby translating changes in mandible position into corresponding changes in the TMJ joint space. This approach enabled the semi-virtual simulation of changes in the TMJ joint space within a common coordinate system.

In the fifth step, the differences between the virtual and semi-virtual models were quantified in terms of mandible positions and simulated changes in the TMJ joint space. 
First, the consistency error was quantified for the target mandible positions and expressed as the rigid-body transformation representing the steps required to align the target mandible positions in the virtual and semi-virtual models. 
Then, the changes in the TMJ joint space were simulated using both models and the results were compared among the models.

\subsection{Ethical approval}

This study was performed in accordance with the principles of the Declaration of Helsinki. 
Approval was granted by the Bioethics Committee of the Silesian Medical Chamber in Katowice, Poland (Resolution No. 33/2021, 22 September 2021). 
All participants (specifically, a single participant) provided their written informed consent to participate in the study. 
All the procedures were performed in accordance with relevant guidelines and regulations.

\subsection{Multimodal data sources}

The presented study is based on multimodal data obtained from a single adult participant (43-year-old male volunteer).
Multimodal data include (1) representations
of maxillofacial anatomical structures acquired using a CBCT scanner (CBCT images), (2) representations of facial morphology during mandibular motion acquired using a facial scanner (Facial images), and (3) digital representations a dental plaster model, replicating tooth and gingival morphology, imaged using an optical scanner (Dental images). Details on the imaging modalities, image types, and their corresponding coordinates in the multimodal dataset are summarized in Table~\ref{tab:imaging-modality}.

\begin{table}[ht]
\centering
\caption{Details on the imaging modalities, image types, and corresponding coordinates in the integrated multimodal dataset}\label{tab:imaging-modality}
\begin{tabular}{
>{\raggedright\arraybackslash}p{0.25\linewidth}
p{0.15\linewidth}
>{\raggedright\arraybackslash}p{0.3\linewidth}
p{0.15\linewidth}
}
\toprule
\textbf{Imaging modality} & \textbf{Image type} & \textbf{Descriptors} & \textbf{Coordinates}\\
\midrule

CBCT scanner (Green X 16, Vatech Co.) & CBCT images & 3D representations
of maxillofacial anatomical structures & \(M_C\)\\
\addlinespace[2mm]
Facial scanner (3dMDface, 3dMD LLC) & Facial images & 3D representations of facial morphology during mandibular motion & \(M_F\) \\
\addlinespace[2mm]
Optical scanner (VIVID 9i, Konica Minolta Sensing) & Dental images & 3D digital representations of a~dental plaster model replicating tooth and gingival morphology & \(M_D\) \\

\bottomrule
\end{tabular}
\end{table}

CBCT images were acquired using a Green X 16 CBCT scanner (Vatech Co., Ltd., Hwaseong-si, Gyeonggi-do, Korea) and the following imaging settings: 87 kVp and 6 mA, with an exposure time of 9 s, a field of view of 160 × 90 mm, and an isotropic voxel size of 0.20 mm.

Facial images were acquired using a 3dMDface facial scanner (3dMD LLC, Atlanta, GA, USA) following the previously described protocol \cite{TomakaG2019, TOMAKA2025110311}. The 3dMDface facial scanner represents a stereophotogrammetric system which uses synchronized modular camera units comprising stereo camera pairs for surface geometry, color cameras for texture acquisition, and projected speckle illumination to facilitate stereo correspondence. 
The 3D surface geometry was reconstructed using active stereophotogrammetric triangulation. The nominal acquisition time was approximately 1.5--2 ms, with a reported geometric accuracy of approximately 0.2 mm. 
A custom acquisition procedure was developed to enable sequential recording of synchronized multi-camera image sets without concurrent 3D reconstruction. This allowed image sets to be acquired at approximately 7 sets per second, with the acquisition rate limited primarily by data transfer to the computer. 
Mandibular motions were captured using two mandibular bows, the first attached to the forehead and the second attached to the mandibular teeth. Thus, mandibular bows were used as a reference to align the position of the facial surface and the teeth. The mandibular motions were then decomposed into a static component (represented by the position of the forehead) and a dynamic component (represented by the position of the mandible). The forehead motions were removed by aligning static facial surfaces across time frames.
The 3D facial surfaces were subsequently reconstructed using the original 3dMD reconstruction software supplied with the scanner and the same reconstruction procedure as for conventional single acquisitions.

Dental images were acquired using a non-contact VIVID 9i optical scanner (Konica Minolta Sensing, Inc., Sakai, Osaka, Japan) following the previously described protocol \cite{TOMAKA2025110311}. The optical scanner, based on laser triangulation, operated with the TELE lens (25 mm focal length), providing a nominal xyz accuracy of $\pm0.05$ mm and a z-direction precision ($\sigma$) of 0.008 mm at a measurement distance of approximately 0.6 m. Each acquisition produced up to 640 $\times$ 480 3D points, with a nominal acquisition time of 2.5 s per scan. The scanning of dental plaster models was performed on a rotary table from eight viewpoints during a single rotation. Each occlusal splint was placed within a dental plaster model four times and scanned. Four dental images obtained in this way represented four replicates of the physical realizations.

Within a single-participant study design, this study serves as a proof-of-concept investigation of the feasibility of multimodal data integration for semi-virtual and virtual simulations of changes in the TMJ joint space. The developed methodology can further be used in the clinical cohort study.

\subsection{Multimodal data integration using transformation tree}\label{sec:transf-tree}

\begin{figure}[!ht]
\centering
\includegraphics[width=\linewidth]{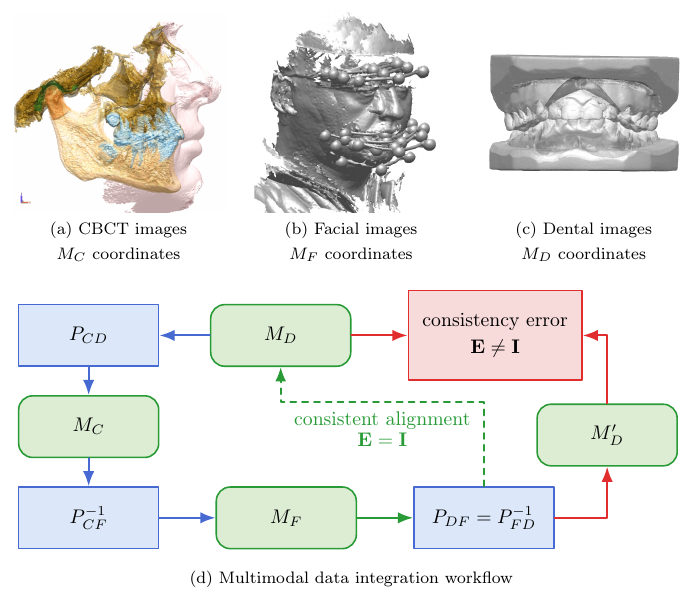}
\caption{Scheme of the multimodal data integration.
(a) CBCT images - segmented anatomical structures represented in the CBCT coordinate system (\(M_C\)).
(b) Facial images - mandibular motions represented in the facial coordinate system (\(M_F\)).
(c) Dental images - digitalized dental plaster models represented in the dental coordinate system (\(M_D\)).
(d) Simplified transformation tree workflow linking the three coordinate systems. \(P_{AB}\) denotes the rigid-body transformation mapping data from coordinate system \(B\) to \(A\).
The alignment consistency is evaluated from the composed transformation \(E=P_{DF}P_{CF}^{-1}P_{CD}\).
The degree of alignment is characterized by the consistency error (\(E\)) and identity (\(I\)). The green path indicates the exact consistent alignment corresponding to \(E=I\). The red path indicates a non-identity consistency error, expressed as transformation, represents the steps required to align the target mandible positions in the virtual (\(M_D\)) 
and semi-virtual (\(M_D'\)) models}.

\label{fig:all_registration_groups}
\end{figure}

Multimodal data integration considered three coordinate systems for three image types: (1) CBCT images ($C$) (Figure~\ref{fig:all_registration_groups}(a)), (2) Facial images ($F$) (Figure~\ref{fig:all_registration_groups}(b)), (3) and Dental images ($D$) (Figure~\ref{fig:all_registration_groups}(c)), where data represented in these coordinate systems were denoted by $M_C$, $M_F$, and $M_D$, respectively. Each group of data contained the original images and the meshes or segmented regions used for registration of 3D shapes. The transformation tree, in which the rigid-body transformations relate data acquired in different coordinate systems, was then applied following our previously published framework~\cite{POJDA2025102093,TOMAKA2025110311} (Figure~\ref{fig:all_registration_groups}(d)). 

The registration of 3D shapes, including free-form curves and surfaces, enables the alignment of images acquired from different modalities within a common coordinate system \cite{besl1992method}, thereby obtaining consistent representations of bones, teeth, and soft tissues. 
The alignment was performed using the Iterative Closest Point (ICP) algorithm \cite{besl1992method,rusinkiewicz2001efficient} for the following anatomically corresponding structures: 
(1) the teeth for linking the CBCT--Dental coordinate systems, 
(2) the facial surface (face) and the teeth for linking the CBCT--Facial coordinate systems, 
and (3) the teeth for linking the Dental--Facial coordinate systems. Multimodal data integration was performed using dpVision software.

\subsubsection{Static alignment of 3D shapes}

The alignment of static images was modeled using the rigid-body transformations $\mathbf{P} \mathbf{M}_1$ in common coordinate system:
\[
\mathbf{M}_1' = \mathbf{P} \mathbf{M}_1 =
\begin{bmatrix}
\mathbf{R} & \mathbf{t} \\
0 & 1
\end{bmatrix} \mathbf{M}_1,
\]
where $\mathbf{R} \in SO(3)$ is a rotation matrix and $\mathbf{t} \in \mathbb{R}^3$ is a translation vector.

Transformations between coordinate systems were estimated using the ICP algorithm applied to the corresponding regions of interest:
\[
\begin{split}
\mathbf{P}_{CD} &= \operatorname{ICP}(\mathbf{M}_C^{\mathrm{teeth}}, \mathbf{M}_D^{\mathrm{teeth}}) \\
\mathbf{P}_{CF} &= \operatorname{ICP}(\mathbf{M}_C^{\mathrm{face.teeth}}, \mathbf{M}_F^{\mathrm{face.teeth}}) \\
\mathbf{P}_{FD} &= \operatorname{ICP}(\mathbf{M}_F^{\mathrm{teeth}}, \mathbf{M}_D^{\mathrm{teeth}})
\end{split}
\]
where $\mathbf{P}_{AB}$ maps the points from the coordinate system $B$ to $A$.

The alignment consistency was evaluated using:
\[
\mathbf{E} = \mathbf{P}_{DF} \mathbf{P}_{CF}^{-1} \mathbf{P}_{CD}, \qquad \mathbf{P}_{DF} = \mathbf{P}_{FD}^{-1},
\]
where consistent alignment corresponding to $\mathbf{E} \approx \mathbf{I}$.

\subsubsection{Dynamic alignment of 3D shapes}

The alignment of dynamic images was modeled using the time-dependent rigid-body transformation $\mathbf{P}_M(t)$ in a common coordinate system:

\[
\Big[\mathbf{M'}_{S}\quad\mathbf{M'}_{M}(t)\Big] =
\mathcal{P}\cdot
\Big[\mathbf{M}_{S}\quad\mathbf{P}_M(t)\cdot\mathbf{M}_{M}\Big],
\]

where $\mathbf{P}_M(t)$ is the movement represented as transformations in $t$ moment of time, and $M_S$, $M_M$ are representations of the static component and the dynamic component \cite{TOMAKA2025110311}.

The mandibular motions defined in the facial coordinate system (\(M_F\)) can be transferred to the CBCT coordinate system (\(M_C\)) and the dental coordinate system (\(M_D\)) using:
\[\begin{split}
\mathbf{P}_{D}(t) &= \mathbf{P}_{FD}^{-1} \cdot \mathbf{P}_{F}(t) \cdot \mathbf{P}_{FD} \\
\mathbf{P}_{C}(t) &= \mathbf{P}_{FC} ^{-1}\cdot \mathbf{P}_{F}(t) \cdot \mathbf{P}_{FC}
\end{split}\]

This enables translation of mandibular motions from \(M_F\) to \(M_C\) and \(M_D\) facilitating simulations of changes in the TMJ joint space under various mandible positions.

\subsection{Mandible positions\label{sec:positioning}}

\begin{figure}[ht]
\centering
    \includegraphics[width=\linewidth]{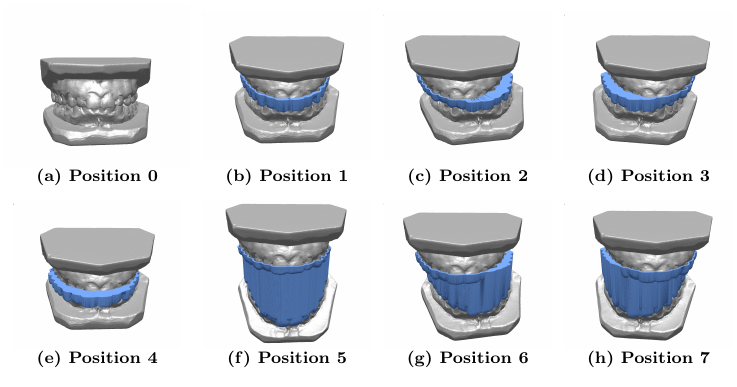}
    \caption{Visualization of the mandible positions considered in the virtual model and in the semi-virtual model. 
    (a) Position 0 - the neutral position. 
    (b) Position 1 - the mild mouth opening.
    (c) Position 2 - the mild mouth opening with left lateral deviation.
    (d) Position 3 - the mild mouth opening with right lateral deviation.
    (e) Position 4 - the mild mouth opening with mandible protrusion.
    (f) Position 5 - the moderate mouth opening.
    (g) Position 6 - the moderate mouth opening with left lateral deviation.
    (h) Position 7 - the moderate mouth opening with right lateral deviation.}
    \label{fig:mandible_positions}
\end{figure}

During facial scanning, the participant was instructed to perform mandibular motions within the physiological range. Image acquisition began with the mandible in the neutral position (position 0), with the mouth closed and the teeth in habitual occlusion. The participant then performed the following movements in sequence: mild mouth opening (position 1), mild mouth opening with left lateral deviation (position 2), mild mouth opening with right lateral deviation (position 3), mild mouth opening with mandible protrusion (position 4), moderate mouth opening (position 5), moderate mouth opening with left lateral deviation (position 6), and moderate mouth opening with right lateral deviation (position 7). In total, eight mandible positions were acquired and selected for subsequent modeling. The characteristics of these positions are summarized in Table~\ref{tab:mandible-positions} and illustrated in Figure~\ref{fig:mandible_positions}.

\begin{table}[ht]
\centering
\caption{Mandible positions considered in the virtual model and in the semi-virtual model including the neutral position (0) and each target position (1-7). The reported distances for depression, protrusion, and lateral deviation were approximated at the level of the incisors, between the mandibular and maxillary incisors along the midline.}\label{tab:mandible-positions}
\begin{tabular}{
>{\raggedright\arraybackslash}p{0.15\linewidth}
p{0.25\linewidth}
p{0.15\linewidth}
p{0.15\linewidth}
p{0.2\linewidth}
}
\toprule
\textbf{Numbers} & \textbf{Descriptors} & \textbf{Depression} & \textbf{Protrusion} & \textbf{Lateral deviation}\\
\midrule
Position 0 & Neutral position & No & No & No\\
\addlinespace[2mm]
Position 1 & Mild mouth opening & 1 cm & No & No \\
\addlinespace[2mm]
Position 2 & Mild mouth opening with left lateral deviation & 1 cm & No & Left, 1 cm \\
\addlinespace[2mm]
Position 3 & Mild mouth opening with right lateral deviation & 1 cm & No & Right, 1 cm \\
\addlinespace[2mm]
Position 4 & Mild mouth opening with mandible protrusion & 1 cm & 1 cm & No \\
\addlinespace[2mm]
Position 5 & Moderate mouth opening & 3 cm & No & No \\
\addlinespace[2mm]
Position 6 & Moderate mouth opening with left lateral deviation & 2 cm & No & Left, 1 cm \\
\addlinespace[2mm]
Position 7 & Moderate mouth opening with right lateral deviation & 2 cm & No & Right, 1 cm \\
\bottomrule
\end{tabular}
\end{table}

The selected mandible positions (0-7) were represented using the rigid-body transformations:

\[
\mathbf{T} =
\begin{bmatrix}
\mathbf{R} & \mathbf{t} \\
\mathbf{0} & 1
\end{bmatrix},
\]

where $\mathbf{R} \in SO(3)$ and $\mathbf{t} \in \mathbb{R}^3$. Transformations map the mandible from a reference position (0) to target positions (1–7). Transformation composition was performed using dpVision software.

In a virtual model, these transformations contained the dental coordinates (\(M_D\)) of the planned mandible positions, which were used to simulate related changes in the TMJ joint space. In a semi-virtual model, these predefined transformations were used to design and fabricate seven occlusal splints replicating the distances and orientations of the teeth in the planned mandible positions (1–7). The occlusal splints were fabricated using 3D printing, with their models were generated as meshes in STL format according to the previously described protocol \cite{Pojda2019, TOMAKA2025104527}. The manufacturing process was carried out using a Bambu Lab A1 mini 3D printer (Bambu Lab, Shenzhen, China), operating according to the fused deposition modeling (FDM) principle. The printer has a build volume of 180 × 180 × 180 mm and an all-metal hot end with a standard 0.4 mm stainless steel nozzle. Polylactic acid (PLA) filament with a diameter of 1.75 mm was used as the printing material. The maximum nozzle and build plate temperatures are 300 °C and 80 °C, respectively, with a nominal maximum toolhead speed of 500 mm/s.

Each occlusal splint was then placed individually on a dental plaster model, where the occlusal splints act as a physical realization of the rigid-body transformation. Each physical realization was then imaged using the optical scanner following the previously described protocol \cite{TOMAKA2025110311}. 
Achieved dental coordinates were integrated into the common coordinate system.
In a semi-virtual model, the rigid-body transformations contained the dental coordinates (\(M'_D\)) of the achieved mandible positions, which were used to simulate related changes in the TMJ joint space.

\subsection{Consistency error quantification}

The consistency error was expressed as the rigid-body transformation that represents the steps required to achieve identity (\(I\)) between the target mandible positions in the virtual and semi-virtual models: 

\[
E=T_{\mathrm{error},i} = T_{\mathrm{semi-virtual},i} T_{\mathrm{virtual},i}^{-1}.
\]

where (\(E\)) is the consistency error and (\(T_{\mathrm{error},i})\) is the rigid-body transformation for each realization $i$, representing each target position (1-7) measured in four replicates. The mandible positions are denoted by $T_{\mathrm{virtual},i}$ for the virtual model and $T_{\mathrm{semi-virtual},i}$ for the semi-virtual model. The perfect reproduction of the planned mandible positions as a physical realization (achieved mandible position) corresponds to $E=I$, while deviations from \(I\) indicate the consistency error introduced by occlusal splints.

The consistency error of the mandible positioning was defined by translation ($t$) and rotation ($r$) components. The $t$ and $r$ were analyzed separately. These correspond to changes in distance and orientation of the mandible between the neutral position (0) and the each target position (1-7), respectively.
Translation describes the distances in the x (latero-lateral) direction (\(t_x\)), the y (dorso-ventral) direction (\(t_y\)), and the z (rostro-caudal) direction (\(t_z\)); whereas, rotation describes the orientations along the x axis (\(r_x\)), along the y axis (\(r_y\)), and along the z axis (\(r_z\)). The x, y, z axes were orthogonal.

Since the mandible can be both translated and rotated, the consistency error was also described using two quantities: the total translation error (\(t_{norm}\)) and the total rotation error (\(\theta\)). The \(t_{norm}\) is expressed in millimeters [mm] and indicates by how many mm the mandible has been displaced from its planned position to its achieved position, regardless of the direction of transformation. The $\theta$ indicates by how many degrees orientation of the mandible differs from the planned position to its achieved position, regardless of the axis of rotation. The $r$ was obtained via the logarithmic map $\mathbf{r} = \mathrm{Log}(\mathbf{R})$, as defined for $SO(3)$~\cite{barfoot2017state,lynch2017modern,hall2015lie}, and expressed in axis--angle form $\mathbf{r} = \theta \mathbf{u}$. For reporting purposes, $r$ and $\theta$ were converted from radians to degrees and expressed in degrees [deg]. 

This approach allows the magnitude of the errors to be reported without having to analyze each of the six transformation components separately. Consequently, the distance (\(t_x\), \(t_y\), \(t_z\), \(t_{norm}\)) and orientation (\(r_x\), \(r_y\), \(r_z\), \(\theta\)) in which the planned mandible position differed from the achieved mandible position were quantified and presented for each target mandible position (1-7) and for all combined positions (All) to indicate the systematic offset. The quantification of consistency error, including transformation management, was performed using dpVision software.

\subsection{Simulation of TMJ joint space changes}

The mandibular condyles and the glenoid fossae were semi-automatically segmented from the CBCT images by an experienced orthodontist. The relevant anatomical surfaces were delineated along a border defined by hyperdense areas representing cortical bone. This expert-guided anatomical delineation was performed for the left and right TMJ separately using RapidForm and dpVision software. 
RapidForm software was used extensively during the earlier stages of TMJ segmentation for basic segmentation, manual correction (manual refinement of segmented anatomical surfaces and manual refinement of selected surface-based registration operations), and final alignment 3D shapes of surface meshes. The meshes represented corresponding anatomical regions and their points were sampled at the same resolution as in the previously used protocol \cite{Pojda2019}. Subsequently, dpVision software was used for the edition, visualization, and integration of the reconstructed anatomical structures with the remaining multimodal data within a common coordinate system. 
The segmented anatomical structures were saved as 3D surface meshes.

The rigid-body transformations of the mandible were propagated to the 3D surface meshes of the TMJ using dpVision software.
The position of the mandibular condylar in relation to the glenoid fossa was simulated simultaneously in the left and right TMJ while maintaining participant-specific mandible geometry modeled in the common coordinate system. This provides simulation of the spatial relationships between the articular surface within the TMJ.

The simulation of the TMJ joint space was performed for the neutral position (0) and each target position (1-7) using the coordinates considered in the virtual model and in the semi-virtual model, separately. The resulting positions of the articular surfaces were quantified using distance-based metrics (the joint space distance and the model disagreement). 

\subsection{Joint space distance and model disagreement quantification}

The joint space distance was calculated between the glenoid fossa and the mandibular condyle across the entire 3D surface meshes. The joint space distance was evaluated at each of the seven target positions (1–7) for both the virtual and semi-virtual models, separately for the left and right TMJs.

The joint space distance was visualized using distance maps showing the shortest distances between the meshes. The method used to visualize distance maps of the shortest distances between points in the glenoid fossa and the mandibular condyles is a conceptual extension of the approach presented in~\cite{SHU2022107149}. The reference approach was applied to single points, while in this study, the maps were applied to the entire 3D surface meshes. Visualizations on the distance maps were limited to the joint space distance range of 0 to 10~mm. 

The model disagreement was calculated as the difference between the joint space distances computed in the virtual and semi-virtual models.
For each realization \(i\), the model disagreement was characterized using the absolute mean ($\mu$) and  standard deviation (SD; $\sigma$) ~\cite{Wasserman2004,CasellaBerger2002} calculated using: 

\[
\mu = \overline{|d|} =   \frac{1}{N}\sum_{i=1}^{N}|d_i|,
\]

and

\[
\sigma = \sqrt{\frac{1}{N}\sum_{i=1}^{N}(|d_i|-\overline{|d|})^2}\label{eq:sim-error}
\]

where $N$ is the sum of the points in the four replicates for a given mandible position, and $d_i$ is distance between the points $i$ on the surface of the anatomical regions being under interest (the glenoid fossa to the mandibular condyle at the TMJ level and the mandibular and maxillary incisors in the middle line at the dental level) in the virtual and semi-virtual models. Since model disagreement are absolute, they indicate the magnitude of the disagreement, not its direction. The model disagreement was computed for each target position (1-7) at the TMJ level - for the left TMJ and the right TMJ separately, as well as for both TMJs jointly as a bilateral model disagreement. The model disagreement was also calculated at the dental level. 

For the TMJ level, all points that represent model disagreement fell within the range of 0.0–4.5 mm; therefore, the visualization threshold was set at 4.5 mm. 
For the dental level, all points that represent model disagreement fell within the range of 0.0–2.0 mm; therefore, the visualization threshold was set at 2 mm. 
For both levels, all points on the respective surfaces were included in the analyses.

Given that distance changes can be more informative than absolute values~\cite{YANG2013}, the model disagreement was visualized using distance maps showing the differences in distances between the meshes computed using the virtual model and the semi-virtual model. Visualizations on the distance maps were limited to the model disagreement range of 0 to 4.5~mm. 

Distance-based metric calculation and visualization was performed using auxiliary analysis scripts.

\subsection{Statistical analysis}

The data set contained representations of the mandible and articular surfaces in seven target positions (1-7) measured in four replicates, resulting in 28 realizations $i$. 
First of all, for each realization $i$, the consistency error was described using variability measures (log-residual mean, mean, SD) and covariance measures (variance, share, $r_{95}$, eigenvectors).
Then, for each realization $i$, the simulated changes in the joint space were described using variability measures (mean, SD) and the model disagreement were compared between the main anatomical locations. 
Statistical analysis was performed using auxiliary analysis scripts.
The final model disagreement comparison was performed using GraphPad Prism version 6 (GraphPad Software Inc., San Diego, CA, USA).

The representative consistency error transformation was computed using iterative log-residual averaging on $SE(3)$. Given the input transformations $T_i$, the initial estimate was $M_0=T_1$. At iteration $k$, the update vector was calculated as
\[
\bar{\xi}_k=\frac{1}{N}\sum_{i=1}^{N}\operatorname{Log}(M_k^{-1}T_i),
\qquad
M_{k+1}=M_k\operatorname{Exp}(\bar{\xi}_k),
\]
where each six-dimensional vector is ordered as rotation followed by translation. Iteration stopped when the Euclidean norm of $\bar{\xi}_k$ was below $10^{-9}$ or after 20 updates. This norm was applied directly to the rotation coordinates in radians and translation coordinates in millimeters, without additional scaling; the tolerance is therefore the numerical stopping criterion for this implementation, not a unit-independent distance threshold. Applied to all 28 realizations, the iteration converged after three updates; the per-position means converged after two or three updates. We refer to the result as the iterative $SE(3)$ log-residual mean, rather than claiming that the implementation explicitly minimizes a specified left-invariant Riemannian distance.

For presentation in a common coordinate system, the resulting mean transformation was represented to first order in its rotation vector as:

\[
\mathbf{T}_{\mathrm{mean}} \approx
\begin{bmatrix}
1 & -r_z & r_y & t_x \\
r_z & 1 & -r_x & t_y \\
-r_y & r_x & 1 & t_z \\
0 & 0 & 0 & 1
\end{bmatrix}.
\]

where $t_x$, $t_y$, and $t_z$ denote the translation components, and $r_x$, $r_y$, and $r_z$ are the rotation-vector components. This first-order matrix representation was used only to present the computed mean; the mean itself was obtained using the iterative logarithm-exponential update above. The rotation angles of the 28 error transformations ranged from 0.36$^\circ$ to 4.55$^\circ$ (mean 1.85$^\circ$, median 1.67$^\circ$), and the rotation angle of their overall log-residual mean was 1.429$^\circ$. At this mean angle, the maximum element-wise difference between the first-order and exact rotation matrices was approximately $3.1\times10^{-4}$. The variability measures were presented in Tables for the distance metrics ($t_x$, $t_y$, $t_z$, $t_{norm}$) and the orientation metrics ($r_x$, $r_y$, $r_z$, $\theta$) using descriptive statistics.

Covariance measures were computed using principal-component analysis (PCA) to explore the dominant directions and the dominant orientations of the observed variability of the consistency error. In PCA, all realizations $i$ of the translation vector \((t_x,t_y,t_z)\) and the rotation vector \((r_x,r_y,r_z)\) were combined. Thus, among the 28 realizations, each represents one observation of a 3D random variable. From these realizations, a \(3\times3\) covariance matrix was calculated, which describes both the variability of the individual components ($t$, $r$) and the correlations between them. The PCA was then performed by computing the eigenvectors and eigenvalues of the covariance matrix.
Eigenvectors ($[v_x,v_y,v_z]$) identify the principal directions of variability.
Eigenvalues (variance) quantify the variance along these directions.
Share is the percentage of individual variance in the total variability of a given principal component (PC).
The $r_{95}$ is the semi-axis lengths of the covariance ellipsoid scaled to the nominal 95\% level under a Gaussian approximation.
The first principal component (PC1) represents the direction of maximum variance.
Thus, the covariance matrix describes the overall 3D variability of translation and rotation separately, while PCA identifies its dominant directions and orientations, and how much variance is associated with them. This enables determining whether the consistency errors are approximately isotropic, described by variance that is identical in all spatial directions, or anisotropic, described by variance predominantly distributed along specific spatial directions. Given that the observations are not necessarily independent, the PCA results should be interpreted as descriptive.

For the simulated changes in the joint space, the variability measures were presented in Tables for distance-based metrics (the joint space distance and the model disagreement) using descriptive statistics. Additionally, the absolute difference of the joint space distance was calculated between the left and right TMJs to indicate symmetry or asymmetry of the joint space distance in the target positions. Symmetry was considered when this absolute difference < \(t_{norm}\) calculated for each target position, while asymmetry was considered when this absolute difference > \(t_{norm}\), respectively.

Moreover, the mean model disagreement was calculated as mean and SD for each data series for the dental and the TMJs, specifically for the left TMJ and the right TMJ as unilateral mean model disagreement, as well as for both TMJs as bilateral mean model disagreement. 
The model disagreements were compared between the main anatomical locations that were included in the rigid-body transformations (the dental level and the TMJ level). 
For each anatomical location, model disagreement was presented as a data series comprising individual realizations ($i$). Prior to statistical analysis, the four replicates obtained for each target position were averaged, yielding one mean value per target position. Consequently, each data series consisted of seven averaged realizations ($i_a$), corresponding to seven consecutive target positions (1–7), and thus provided seven paired realizations for each comparison.
Data series were assessed for data distribution using the Shapiro-Wilk normality test. Given that three of four data series did not follow a normal distribution, pairs of data series were compared using two-sided Wilcoxon matched-pairs signed-rank tests. The effect size was quantified using the matched-pairs rank-biserial correlation \cite{kerby2014simple}. The Hodges--Lehmann estimator was calculated to quantify the location of the paired differences, together with its 95\% confidence interval based on rank statistics \cite{hodges1963estimates,bauer1972constructing}. Statistical significance was set at $\alpha = 0.05$, with p-values below 0.05 considered statistically significant.

\section{Results}

\subsection{Variability of consistency error of the mandible positioning}

\begin{table}[t]
\centering
\caption{The consistency error between virtual and semi-virtual models described by variability measures (log-residual mean, mean, standard deviation (SD)) of translation component ($t$) in the rigid-body transformation for the coordinates in a common coordinate system. Measures were calculated for each target mandible position (1-7) and for all combined positions (All) to indicate the systematic offset. Measures of translation are expressed in millimeters [mm].}
\label{tab:uchybM_translation_compact}
\begin{tabular}{clrrrrrrrr}
\hline
$t$ & \multirow{2}{*}{Measures} & \multicolumn{8}{c}{Positions}\\
{[mm]} & & 1 & 2 & 3 & 4 & 5 & 6 & 7 & All\\
\hline
\hline
\multirow{3}{*}{$t_x$} & log-residual mean & -0.91 & -1.61 & -1.92 & -1.99 & -3.67 & -2.83 & -1.84 & -2.11 \\
 & mean & -0.91 & -1.61 & -1.92 & -1.99 & -3.68 & -2.83 & -1.83 & -2.11 \\
 & SD & 0.17 & 1.81 & 0.62 & 1.23 & 2.51 & 2.19 & 1.47 & 1.66 \\
\hline
\multirow{3}{*}{$t_y$} & log-residual mean & -1.34 & -0.41 & 0.31 & -1.13 & -0.66 & -1.16 & 0.76 & -0.51 \\
 & mean & -1.34 & -0.42 & 0.31 & -1.13 & -0.69 & -1.18 & 0.75 & -0.53 \\
 & SD & 0.14 & 0.36 & 0.23 & 0.47 & 0.58 & 0.87 & 0.58 & 0.88 \\
\hline
\multirow{3}{*}{$t_z$} & log-residual mean & -0.48 & 0.05 & 0.20 & -0.49 & -1.08 & -0.82 & -0.84 & -0.50 \\
 & mean & -0.48 & 0.06 & 0.20 & -0.48 & -1.06 & -0.81 & -0.83 & -0.49 \\
 & SD & 0.13 & 0.20 & 0.13 & 0.16 & 1.14 & 0.61 & 0.09 & 0.63 \\
\hline
\hline
\multirow{3}{*}{$t_{norm}$} & log-residual mean & 1.69 & 1.66 & 1.96 & 2.33 & 3.89 & 3.17 & 2.16 & 2.23 \\
 & mean & 1.70 & 1.70 & 1.97 & 2.46 & 4.08 & 3.43 & 2.40 & 2.54 \\
 & SD & 0.11 & 1.80 & 0.63 & 0.97 & 2.42 & 1.90 & 0.98 & 1.55 \\
\hline
\end{tabular}
\begin{tablenotes}[flushleft]
\footnotesize
\item[] \textit{Footnotes:}
\(t_x\) is the translation in the x (latero-lateral) direction, \(t_y\) is the translation in the y (dorso-ventral) direction, \(t_z\) is the translation in the z (rostro-caudal) direction, and \(t_{norm}\) is the total translation error representing the magnitude of the transformation vector.
\end{tablenotes}
\end{table}

\begin{table}[t]
\centering
\caption{The consistency error between virtual and semi-virtual models described by variability measures (log-residual mean, mean, standard deviation (SD)) of rotation component ($r$) in the rigid-body transformation for the coordinates in a common coordinate system. Measures were calculated for each target mandible position (1-7) and for all combined positions (All) to indicate the systematic offset. Measures of rotation are expressed in degrees [deg].}
\begin{tabular}{clrrrrrrrr}
\hline
$r$ & \multirow{2}{*}{Measures} & \multicolumn{8}{c}{Positions}\\
{[deg]} & & 1 & 2 & 3 & 4 & 5 & 6 & 7 & All\\
\hline
\hline
\multirow{3}{*}{$r_x$} & log-residual mean & -0.64 & 0.03 & 0.66 & -0.54 & -0.80 & -0.48 & 0.13 & -0.24 \\
 & mean & -0.64 & 0.03 & 0.66 & -0.54 & -0.80 & -0.48 & 0.13 & -0.24 \\
 & SD & 0.14 & 0.18 & 0.17 & 0.40 & 0.85 & 0.48 & 0.33 & 0.62 \\
\hline
\multirow{3}{*}{$r_y$} & log-residual mean & 1.38 & 0.62 & 1.32 & 0.56 & 1.37 & 0.47 & 0.02 & 0.82 \\
 & mean & 1.38 & 0.62 & 1.32 & 0.56 & 1.37 & 0.47 & 0.02 & 0.82 \\
 & SD & 0.14 & 1.05 & 0.43 & 0.55 & 0.92 & 1.08 & 0.60 & 0.84 \\
\hline
\multirow{3}{*}{$r_z$} & log-residual mean & -0.64 & 0.93 & 1.11 & 1.35 & 1.81 & 2.34 & 1.11 & 1.15 \\
 & mean & -0.64 & 0.93 & 1.11 & 1.35 & 1.81 & 2.34 & 1.11 & 1.15 \\
 & SD & 0.11 & 0.81 & 0.28 & 0.54 & 1.35 & 1.33 & 0.85 & 1.16 \\
\hline
\hline
\multirow{3}{*}{$\theta$} & log-residual mean & 1.65 & 1.12 & 1.85 & 1.56 & 2.41 & 2.44 & 1.12 & 1.43 \\
 & mean & 1.66 & 1.20 & 1.86 & 1.66 & 2.57 & 2.67 & 1.33 & 1.85 \\
 & SD & 0.03 & 1.24 & 0.50 & 0.57 & 1.53 & 1.25 & 0.70 & 1.01 \\
\hline
\end{tabular}
\label{tab:uchybM_rotation_compact}
\begin{tablenotes}[flushleft]
\footnotesize
\item[] \textit{Footnotes:}
\(r_x\) is the rotation along the x (latero-lateral) axis, \(r_y\) is the rotation along the y (dorso-ventral) axis, \(r_z\) is the rotation along the z (rostro-caudal) axis, and $\theta$ is the total rotation error representing the rotation angle of the transformation vector.
\end{tablenotes}
\end{table}

Within the analyzed dataset, \(T_{mean}\) showed a systematic offset in the $t$, dominated by translation in the $x$ direction (log-residual mean: $-2.11$~mm), with less pronounced contributions in the $y$ (log-residual mean: $-0.51$~mm) and $z$ (log-residual mean: $-0.50$~mm) directions. This offset was repeatable for positions 2-7 and was demonstrated in Table \ref{tab:uchybM_translation_compact}. The \(t_{norm}\) of the log-residual mean ranged from $1.66$~mm to $3.89$~mm, with a descriptive mean of $2.23$~mm, indicating by how many mm has the mandible been displaced from its planned position (in the virtual model) to its achieved position (in the semi-virtual model), regardless of the direction of transformation.

Moreover, \(T_{mean}\) showed a systematic offset in the $r$, dominated by rotation along the $z$ axis (log-residual mean: $1.15$~deg), with smaller contributions in the $x$ (log-residual mean: $-0.24$~deg) and $y$ (log-residual mean: $0.82$~deg) axes. This offset was repeatable for positions 2 and 4-7, and was demonstrated in Table \ref{tab:uchybM_rotation_compact}. The $\theta$ of the log-residual mean ranged from $1.12$~deg to $2.44$~deg, with a descriptive mean of $1.43$~deg, indicating by how many degrees the mandible orientation differs from the planned position (in the virtual model) to its achieved position (in the semi-virtual model), regardless of the axis of rotation. Consequently, one may observe that the variability measures for the consistency error for the $t$ dominated over the $r$ of the rigid-body transformation.

\subsection{Covariance of consistency error of the mandible positioning}

\begin{table*}[t]
\centering
\caption{
The consistency error between virtual and semi-virtual models described by covariance measures (variance, share, $r_{95}$) using the Principal Component Analysis (PCA) for translation ($t$) and rotation ($r$) components in the rigid-body transformation. Measures were calculated for all positions combined to indicate the systematic offset.}
\label{tab:pca}
\begin{threeparttable}
\begin{tabular}{ccrrrrrr}
\hline
Data space & PC & Variance & Share & $r_{95}$ & \multicolumn{3}{c}{Eigenvectors} \\
 & & & {[\%]} & & $v_x$ & $v_y$ & $v_z$ \\[1mm]
\hline
$t$ & PC1 & 2.63 & 70.30 & 4.53 & -0.99 & -0.15 & -0.04 \\
$[mm]$      & PC2 & 0.76 & 20.37 & 2.44 & -0.16 & 0.96 & 0.24 \\
            & PC3 & 0.35 & 9.33 & 1.65 & -0.01 & -0.25 & 0.97 \\
\hline
$r$ & PC1 & 1.46 & 59.89 & 3.38 & -0.02 & 0.32 & 0.95 \\
$[deg]$  & PC2 & 0.60 & 24.77 & 2.17 & -0.24 & 0.92 & -0.32 \\
         & PC3 & 0.37 & 15.35 & 1.71 & 0.97 & 0.24 & -0.06 \\
\hline
\end{tabular}
\begin{tablenotes}[flushleft]
\footnotesize
\item[] \textit{Footnotes:}
$PC$ is the principal component, variance (eigenvalue) is the amount of information explained by a given PC, share is the percentage of individual variance in the total variability of a given PC, $r_{95}$ is the semi-axis lengths of the covariance ellipsoid scaled to the nominal 95\% level under a Gaussian approximation, eigenvectors determine the directions of new axes (PC) in the data space and are given in the form $[v_x,v_y,v_z]$ for the x (latero-lateral), y (dorso-ventral), and z (rostro-caudal) coordinates.
\end{tablenotes}
\end{threeparttable}
\end{table*}

\begin{figure}[t]
    \centering
    \includegraphics[width=\linewidth]{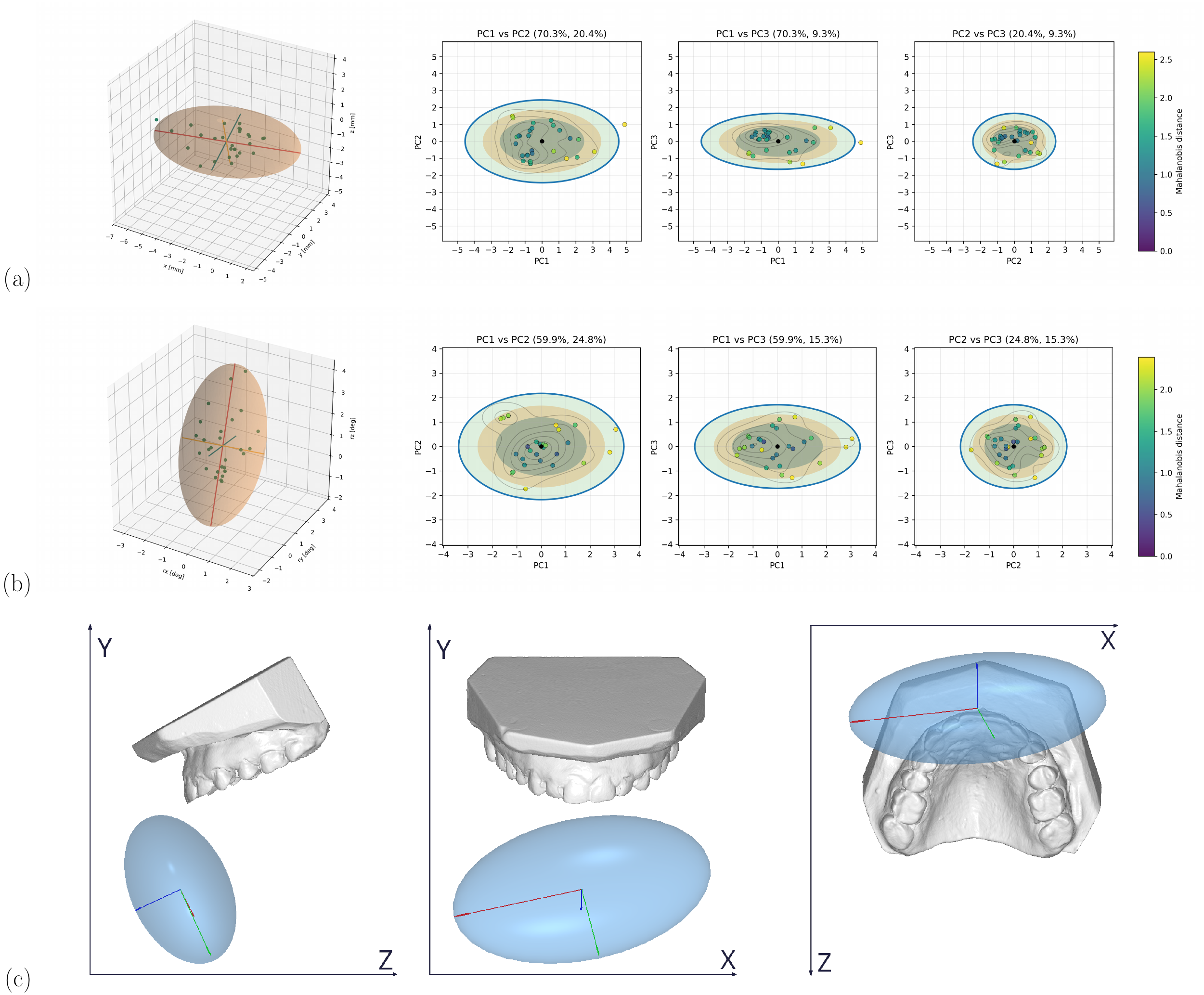}
    \caption{Covariance ellipsoid in the Principal Component Analysis (PCA) for (a) translation ($t$) component, (b) for rotation ($r$) component, and (c) for both  components against the global coordinate system shown together with a dental plaster model of the maxilla. (a, b) Point colors represent Mahalanobis distance and contour lines indicate empirical density. (a-c) The xyz represent direction/axis of the rigid-body transformations for the x (latero-lateral), y (dorso-ventral), and z (rostro-caudal) coordinates.}
    \label{fig:A}
\end{figure}

The dominant directions and the dominant orientations of the observed variability of the consistency error were summarized in Table \ref{tab:pca} to describe the spatial directions along which the differences between the virtual and semi-virtual models were greatest. The covariance measures indicated a strong anisotropy of the consistency error defined by $t$. The PC1 explained the $70.30\%$ variance of the translation. This showed that the variability of $t$ is concentrated in a single dominant direction, indicated as the $x$ direction by the eigenvector $v_x$. On the other hand, covariance measures indicated a moderate isotropy of the consistency error defined by $r$. The PC1 explained the $59.89\%$ variance of the rotation, indicating the spatial distribution being more uniform in all directions. However, the eigenvector $v_z$ still indicated some concentration of variability of $r$ along the $z$ axis.

The dominant directions and the dominant orientations of the variability of consistency error were then visualized in Figure \ref{fig:A} as projections onto the PCA planes. The translation and rotation distributions were non-Gaussian and right-skewed; therefore, covariance ellipsoids were used as descriptive second-order representations of the observed systematic offset. Thus, one may observe that in the space of translation vectors, the variability distribution remained $x$ direction concentrated (Figure \ref{fig:A}(a)), while in the space of rotation vectors, it was more uniform in all spatial directions (Figure \ref{fig:A}(b)). 
When $t$ and $r$ were combined with the global coordinate system, it is easy to see that the principal eigenvector aligns with the $x$ direction indicating lateral displacement of the mandible (Figure \ref{fig:A}(c)). This confirmed that translation variability was dominated, while rotation variability remained limited; and indicated the dominant anisotropic latero-lateral direction of the observed variability of the consistency error.

\subsection{Simulation of the TMJ joint space changes}

The joint space distance was simulated using virtual and semi-virtual models, and the results are presented in Table \ref{tab:tmj-distance-models}. One may observe that the mean and SD values are not identical, neither for individual mandible positions nor for the same mandible positions when simulated using different models. In the neutral position, the joint space distance was 2.37 mm for the left TMJ and 2.56 mm for the right TMJ. When the mouth was mildly open (position 1), the joint space distance ranged from 6.28 mm to 7.35 mm in the virtual model and from 7.04 mm to 7.85 mm in the semi-virtual model. Similarly, when the mouth was moderately open (position 5), the joint space distance ranged from 13.03 mm to 13.62 mm in the virtual model and from 11.91 mm to 14.85 mm in the semi-virtual model. This indicates a bilateral symmetrical change in joint space distance in positions involving only the mouth opening.

In contrast, when the mandible was deviated on the left side, the joint space distance of the left TMJ ranged from 5.28 mm to 5.55 mm and the joint space distance of the right TMJ ranged from 10.03 mm to 11.28 mm when the mouth was mildly open (position 2). Moreover, when the mouth was moderately open (position 6), the joint space distance of the left TMJ ranged from 5.20 mm to 5.58 mm and the joint space distance of the right TMJ ranged from 11.48 mm to 14.07 mm. Similarly, but in the opposite direction, when the mandible was deviated on the right side, the joint space distance of the left TMJ ranged from 9.49 mm to 10.64 mm and the joint space distance of the right TMJ ranged from 6.27 mm to 6.86 mm when the mouth was mildly open (position 3). Moreover, when the mouth was moderately open (position 7), the joint space distance of the left TMJ ranged from 9.93 mm to 11.31 mm and the joint space distance of the right TMJ ranged from 6.66 mm to 7.42 mm. This indicates a bilateral asymmetrical change in joint space distance in positions involving the mouth opening with lateral deviation.

\begin{figure}[!h]
\centering
    \includegraphics[width=\linewidth]{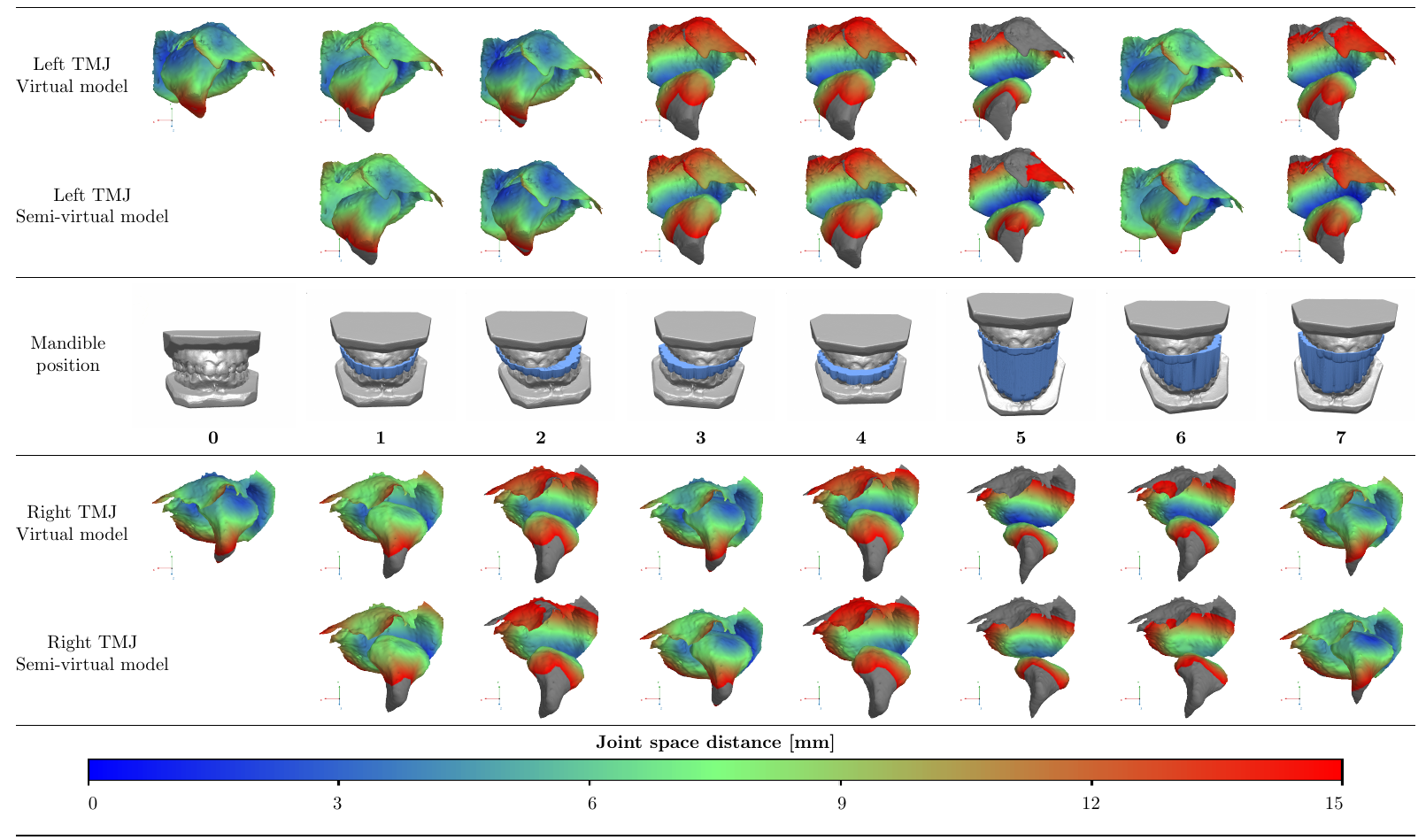}
    \caption{Distance maps of the joint space distance of the left and right temporomandibular joint (TMJ) for each target mandible position (1-7). The joint space distance was simulated using virtual and semi-virtual models. The joint space distance is expressed in millimeters [mm] and visualized using a color scale ranging from 0 mm (blue; indicating narrowing of the simulated joint space) to 15 mm (red; indicating widening of the simulated joint space) on 3D views of the mandibular condyle and the glenoid fossa. The xyz represent direction/axis of the rigid-body transformations for the x (latero-lateral), y (dorso-ventral), and z (rostro-caudal) coordinates.}
    \label{fig:stawy}
\end{figure}

Finally, when the mouth was mildly open and the mandible was protruded (position 4), the mean joint space distances were 10.40 mm in the left TMJ and 9.56 mm in the right TMJ in the virtual model, and 9.94 mm in the left TMJ and 11.49 mm in the right TMJ in the semi-virtual model. This indicates that a bilateral change in joint space distance in positions involving the mandible protrusion may occur symmetrical; however, with a simultaneous slight lateral deviation. These findings demonstrated that changes in mandible positions can be propagated to the articular surface positions using rigid-body transformations, although the results are influenced by the simulation model applied. Regional changes in joint space distance were visualized using the distance maps shown in Figure \ref{fig:stawy}, where a reduction in the simulated joint space (narrowing) is represented by shifts in the distance maps toward blue, whereas an increase in the simulated joint space distance (widening) is represented by shifts toward red.

\subsection{Model disagreement of the TMJ joint space changes}

The model disagreement was calculated as the difference in the joint space distance between the virtual and semi-virtual models, and the results are presented in Table~\ref{tab:tmj-distance-error}. 
One may observe that the mean and SD values of this metric are also not identical for individual mandible positions.

\begin{figure}[!h]
\centering
    \includegraphics[width=\linewidth]{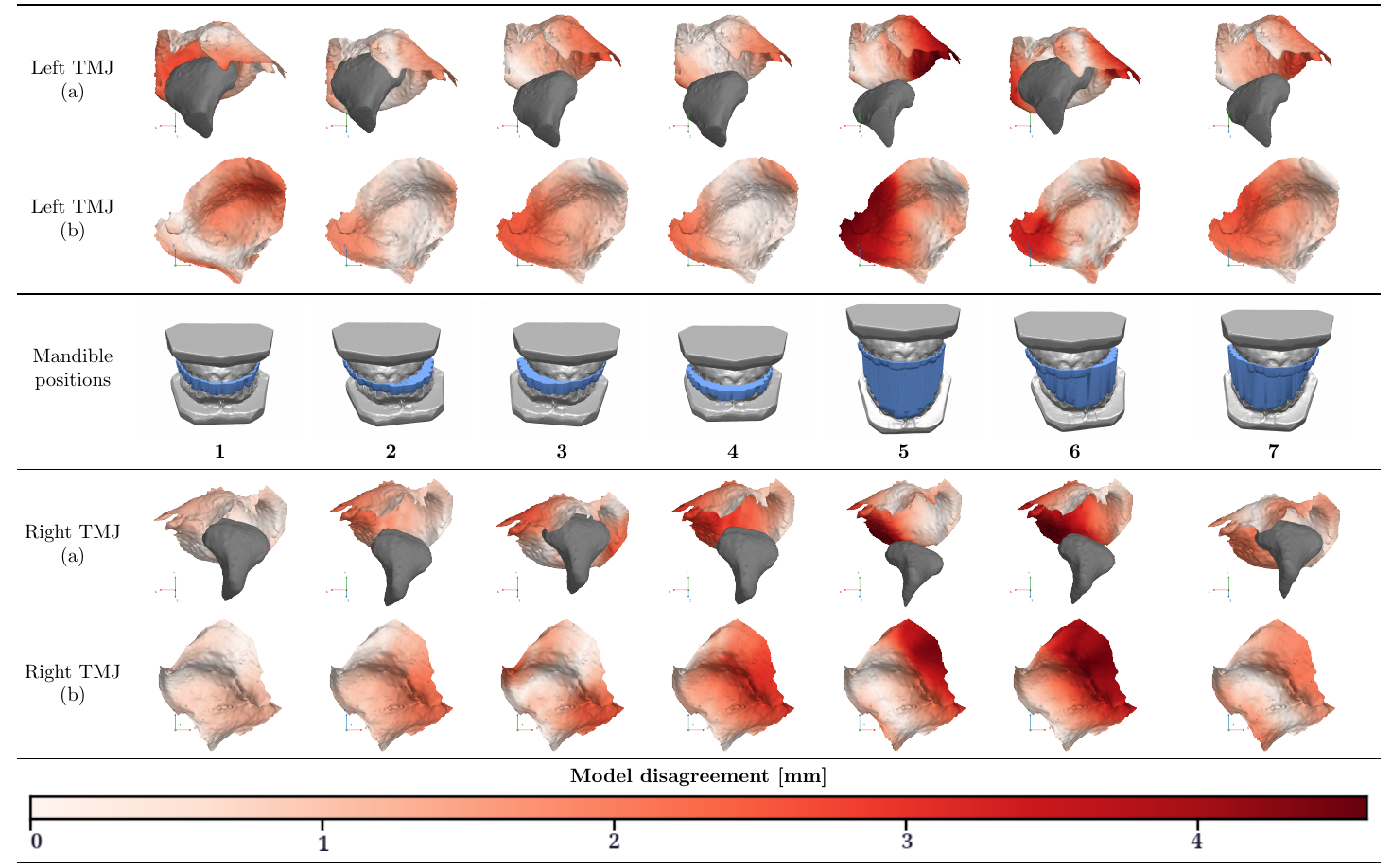}
    \caption{Distance maps of the model disagreement at the temporomandibular joint (TMJ) level for each target mandible position (1-7). The model disagreement was calculated between virtual and semi-virtual models. The model disagreement is visualized as (a) 3D views of the mandibular condyle and the glenoid fossa and (b) an isolated space between the mandibular condyle and the glenoid fossa. The model disagreement is expressed in millimeters [mm] and visualized using a color scale ranging from 0.0 mm (white; indicating identity between models) to 4.5 mm (red; indicating the highest discrepancy between models). The xyz represent direction/axis of the rigid-body transformations for the x (latero-lateral), y (dorso-ventral), and z (rostro-caudal) coordinates.}
    \label{fig:stawy-diff}
\end{figure}

\begin{figure}[!h]
\centering
    \includegraphics[width=\linewidth]{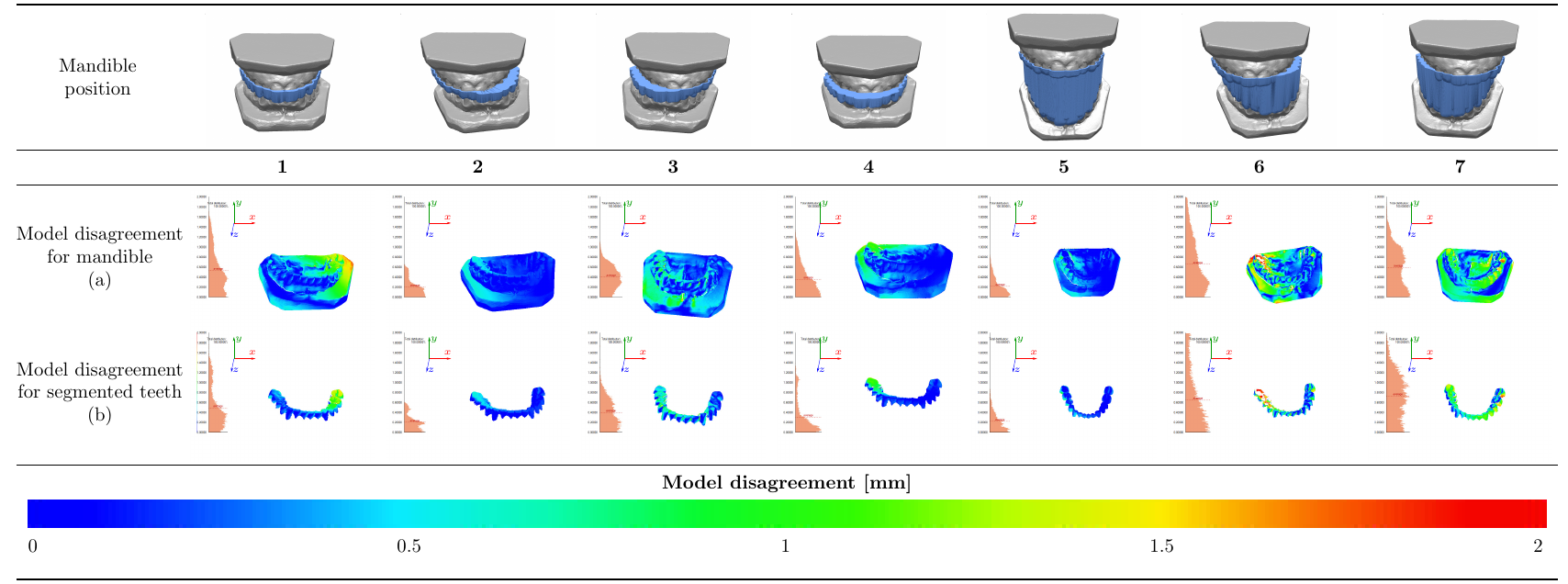}
    \caption{Distance maps of the model disagreement at the dental level for each target mandible position (1-7). The model disagreement calculated between virtual and semi-virtual models. The model disagreement is visualized as 3D views for (a) the mandible and (b) the segmented teeth. The model disagreement is expressed in millimeters [mm] and visualized using a color scale ranging from 0.0 mm (blue; indicating identity between models) to 2.0 mm (red; indicating the highest discrepancy between models). The xyz represent direction/axis of the rigid-body transformations for the x (latero-lateral), y (dorso-ventral), and z (rostro-caudal) coordinates.}
    \label{fig:teeth-diff}
\end{figure}

When the mouth was mildly open (position 1), the model disagreement ranged from 0.68 mm in the right TMJ to 1.33 mm in the left TMJ. Similarly, when the mouth was moderately open (position 5), the model disagreement ranged from 1.83 mm to 1.94 mm in the TMJs on the respective sides. In contrast, when the mandible was deviated on the left side, the model disagreement of the left TMJ ranged from 0.81 mm to 1.49 mm and the joint space distance of the right TMJ ranged from 1.25 mm to 2.77 mm when the mouth was mildly open (position 2) and moderately open (position 6), respectively. Conversely, when the mandible was deviated on the right side, the model disagreement of the left TMJ ranged from 1.22 mm to 1.42 mm and the joint space distance of the right TMJ ranged from 1.27 mm to 1.35 mm when the mouth was mildly open (position 3) and moderately open (position 7), respectively. Moreover, when the mouth was mildly open and the mandible was protruded (position 4), the model disagreement ranged from 0.85 mm in the left TMJ to 1.94 mm in the right TMJ. These findings indicated that because the joint space distance is influenced by the simulation model used, the model-dependent effect can be quantified. However, the resulting model disagreement was not uniform across target mandible positions. Regional variations in model disagreement were visualized using the distance maps and 3D views shown in Figure \ref{fig:stawy-diff} at the TMJ level and in Figure \ref{fig:teeth-diff} at the dental level. For the TMJ level, each value of the model disagreement $ \in [0.0, 4.5]$ mm and the identity between models was labeled in white, while the highest discrepancy between models was labeled in red. 
For the dental level, each value of the model disagreement $ \in [0.0, 2.0]$ mm and the identity between models was labeled in blue, while the highest discrepancy between models was labeled in red. 
The regions in which joint space distance change when transforming from the virtual model to the semi-virtual model were also spatially non-uniform. Therefore, transformation discrepancies can affect different regions of the joint to varying degrees.

The model disagreements were then compared between the main anatomical locations that were included in the rigid-body transformations as shown in Table \ref{tab:comparisons} and Table \ref{tab:wilcoxon_results}. The mean model disagreement was lower (p = 0.0156) at the dental level (0.47 mm) than at the TMJ level - both left (1.28 mm), right (1.60 mm), and bilateral (1.46 mm). However, no differences were found between the TMJs on the respective sides, compared between left and right (p = 0.2969) as well as compared with bilateral measures (p = 0.2969). These findings demonstrated that, in this single-participant case, small transformation discrepancies in mandible position resulted in significantly higher changes at the TMJ level. 

\begin{table}[t]
\centering
\caption{The joint space distance representing distance (mean, standard deviation (SD)) between the mandibular condyle and the glenoid fossa of the left and right temporomandibular joint (TMJ) for each target mandible position (1-7). 
The joint space distance was simulated using (a) virtual and (b) semi-virtual models. The joint space distance is expressed in millimeters [mm]. 
The absolute difference (Difference) between the left and right TMJs was calculated to indicate symmetry (Sym) and asymmetry (Asym) of the joint space distance. Sym was considered when Difference < the total translation error (\(t_{norm}\)), while Asym when Difference > \(t_{norm}\).}
\label{tab:tmj-distance-models}
\begin{tabular}{clrrrrrrr}
\hline
Side & Measures & \multicolumn{7}{c}{Positions} \\
& {[mm]} & 1 & 2 & 3 & 4 & 5 & 6 & 7 \\
\hline
\multirow{2}{*}{Left} & mean & 6.28 & 5.55 & 10.64 & 10.40 & 13.62 & 5.58 & 11.31 \\
& SD & 1.81 & 2.08 & 3.47 & 3.68 & 5.20 & 2.12 & 4.25 \\
\hline
\multirow{2}{*}{Right} & mean & 7.35 & 10.03 & 6.27 & 9.56 & 13.03 & 11.48 & 6.66 \\
& SD & 2.51 & 3.29 & 3.01 & 3.33 & 5.53 & 4.41 & 2.63 \\
\hline
\multirow{2}{*}{Difference} & mean & 1.07 & 4.48 & 4.38 & 0.85 & 0.60 & 5.89 & 4.64 \\
& symmetry & Sym & Asym & Asym & Sym & Sym & Asym & Asym \\
\hline
\end{tabular}
\begin{tablenotes}[flushleft]
\footnotesize
\item[] \textit{Footnotes:}
(a) The joint space distance simulated using a virtual model.
\end{tablenotes}
\vspace{5mm}
\begin{tabular}{clrrrrrrr}
\hline
Side & Measures & \multicolumn{7}{c}{Positions} \\
& {[mm]} & 1 & 2 & 3 & 4 & 5 & 6 & 7 \\
\hline
\multirow{2}{*}{Left} & mean & 7.04 & 5.28 & 9.49 & 9.94 & 11.91 & 5.20 & 9.93 \\
& SD & 1.45 & 2.27 & 3.33 & 3.51 & 4.98 & 2.45 & 4.33 \\
\hline
\multirow{2}{*}{Right} & mean & 7.85 & 11.28 & 6.86 & 11.49 & 14.85 & 14.07 & 7.42 \\
& SD & 2.66 & 3.56 & 3.03 & 3.28 & 5.31 & 4.32 & 3.01 \\
\hline
\multirow{2}{*}{Difference} & mean & 0.81 & 6.00 & 2.62 & 1.55 & 2.94 & 8.87 & 2.51 \\
& symmetry & Sym & Asym & Asym & Sym & Sym & Asym & Asym \\
\hline
\end{tabular}
\begin{tablenotes}[flushleft]
\footnotesize
\item[] \textit{Footnotes:}
(b) The joint space distance simulated using a semi-virtual model.
\end{tablenotes}
\end{table}

\begin{table}[t]
\centering
\caption{The model disagreement representing difference in joint space distance (mean, standard deviation (SD)) between the virtual and semi-virtual models in the left and right temporomandibular joint (TMJ) for each target mandible position (1-7). The model disagreement is expressed in millimeters [mm].}
\label{tab:tmj-distance-error}
\begin{tabular}{clrrrrrrr}
\hline
Side & Measures & \multicolumn{7}{c}{Positions} \\
& {[mm]} & 1 & 2 & 3 & 4 & 5 & 6 & 7 \\
\hline
\multirow{2}{*}{Left} & mean & 1.33 & 0.81 & 1.22 & 0.85 & 1.83 & 1.49 & 1.42 \\
& SD & 0.73 & 1.01 & 0.75 & 0.85 & 1.66 & 1.29 & 0.87 \\
\hline
\multirow{2}{*}{Right} & mean & 0.68 & 1.25 & 1.27 & 1.94 & 1.94 & 2.77 & 1.35 \\
& SD & 0.37 & 1.33 & 0.77 & 0.93 & 1.71 & 1.97 & 1.00 \\
\hline
\end{tabular}
\end{table}

\begin{table}[!h]
\centering
\caption{The mean model disagreement (mean, standard deviation (SD)) compared between the main anatomical locations (the dental level, the temporomandibular joint (TMJ) level). The mean model disagreement is expressed in [mm].}
\label{tab:comparisons}
\begin{tabular}{lccccccc}
\toprule
Region & Side & \multicolumn{2}{c}{Measures [mm]} & \multicolumn{4}{c}{Comparisons [p-value]} \\
& & Mean & SD & Dentition & Left TMJ & Right TMJ & Both TMJs\\
\midrule
Dentition & & 0.47 & 0.11 & - & \textbf{0.0156} & \textbf{0.0156} & \textbf{0.0156} \\
TMJ & Left & 1.28 & 0.36 & \textbf{0.0156} & - & 0.2969 & 0.2969 \\
TMJ & Right & 1.60 & 0.68 & \textbf{0.0156} & 0.2969 & - & 0.2969 \\
TMJs & Both & 1.46 & 0.45 & \textbf{0.0156} & 0.2969 & 0.2969 & - \\
\bottomrule
\end{tabular}
\begin{tablenotes}[flushleft]
\footnotesize
\item[] \textit{Footnotes:} 
Statistical significance was set at \(\alpha = 0.05\); significant p-values are highlighted in bold.
\end{tablenotes}
\end{table}

\begin{table}[!h]
\centering
\caption{The effect size measures and difference estimators for the paired Wilcoxon signed-rank tests comparing the mean model disagreement between the main anatomical locations (the dental level, the temporomandibular joint (TMJ) level).}
\label{tab:wilcoxon_results}
\begin{tabular}{l l c c c c c}
\hline
Region 1 & Region 2 & $n$ & $W$ & $p$-value &
$r_{\mathrm{rb}}$ & HL estimate (95\% CI) [mm] \\
\hline
Dentition & Left TMJ  & 7 & 0 & 0.0156 & -1.00 & $-0.795$ ($-1.339$, $-0.446$) \\
Dentition & Right TMJ & 7 & 0 & 0.0156 & -1.00 & $-1.123$ ($-2.317$, $-0.238$) \\
Dentition & Both TMJs & 7 & 0 & 0.0156 & -1.00 & $-0.946$ ($-1.742$, $-0.527$) \\
Left TMJ & Right TMJ & 7 & 7 & 0.2969 & -0.50 & $-0.262$ ($-1.286$, $0.647$) \\
Left TMJ & Both TMJs & 7 & 7 & 0.2969 & -0.50 & $-0.145$ ($-0.711$, $0.358$) \\
Right TMJ & Both TMJs & 7 & 7 & 0.2969 & 0.50 & $0.117$ ($-0.289$, $0.575$) \\
\hline
\end{tabular}

\footnotesize
\textit{Note:} $n$ denotes the number of paired position-level observations;
$W$ is the Wilcoxon signed-rank statistic, defined as
$\min(W_+,W_-)$; $r_{\mathrm{rb}}$ is the matched-pairs
rank-biserial correlation; HL denotes the Hodges--Lehmann
estimate of the paired difference, with its 95\% confidence
interval (CI). Paired differences were calculated as
Region 1 minus Region 2. Exact two-sided $p$-values were
obtained by enumerating all possible sign assignments of
the observed paired differences.
\end{table}

\section{Discussion}

The present study demonstrates a multimodal data integration methodology for the digital modeling of changes in the TMJ joint space at various mandible positions. 
The proposed methodology effectively integrates CBCT images representing maxillofacial anatomical structures \cite{kaasalainen2021dental}, facial images capturing facial morphology during mandibular motion \cite{heike20103d,shuto2025facial}, and dental images depicting tooth and gingival morphology \cite{keating2008comparison}. This integration within a common spatial coordinate system enables the propagation of mandible positions to both virtual and semi-virtual TMJ models, as well as the quantification of model-based uncertainty.
Given the potential of integrated multimodal data to improve the analysis of the entire stomatognathic system~\cite{TomakaG2019,mangano2018combining,park2022creation}, the proposed approach provides detailed information on the spatial geometric relationships between the articular surfaces of the TMJs. However, its ability to directly represent contact forces \cite{chen1994fea}, pressure distributions \cite{lai2020finite}, and detailed biomechanics of joint movement \cite{chen1998stress} remains limited.

The proposed methodology assumes the rigid-body transformation which is an approach commonly adopted in studies of mandibular motion \cite{li2025integrated,Pojda2019,TOMAKA2025104527}. In this approach, mandibular motions are represented as a sequence of rigid-body transformations, each combining rotation and translation to describe an object's orientation and position in space. The methodology was subsequently extended through the application of a transformation tree framework \cite{POJDA2025102093,TOMAKA2025110311}, which represents hierarchical relationships between coordinate systems and stores successive rigid-body transformations \cite{Hallpike/img.14.6.140455,POJDA2025102093}. Previous studies have demonstrated that transformation trees facilitate the integration of multimodal data across coordinate systems, enabling mandibular motion mapping, comparison of mandible positions, and quantification of positioning errors \cite{TOMAKA2025104527,POJDA2025102093}. The present study further demonstrates that transformation trees also facilitate the simulation, quantification, and comparison of the articular surface positions of the TMJ associated with mandibular motion. The proposed methodology demonstrated, in a single participant, that mandible positions involving mouth opening alone resulted in bilateral, symmetrical changes in joint space distances in the TMJs. In contrast, mandible positions involving mouth opening combined with lateral deviation resulted in bilateral, asymmetrical changes in joint space distances. Furthermore, mandible protrusion may occur concurrently with a slight lateral deviation. These findings, obtained through rigid-body simulation, are in line with the main directions of changes in TMJ joint space distances observed during physiological mandibular motions \cite{SHU2022107149,li2025integrated}. However, to the best of our knowledge, this study is the first to describe these findings using the proposed transformation tree framework. The findings clearly demonstrate that changes in the mandible position can be propagated to the corresponding articular surface positions using rigid-body transformations, although the results depend on the simulation model used.

In the present study, mandible positions within the range of physiological mandibular motions \cite{SHU2022107149,li2025integrated} were modeled using two proprietary approaches: a virtual model based solely on the proposed transformation tree framework \cite{POJDA2025102093,TOMAKA2025110311} and a semi-virtual model combining the same transformation tree framework \cite{POJDA2025102093,TOMAKA2025110311} with physical reproduction of mandible positions using dental plaster models \cite{TOMAKA2025110311}. To the best of our knowledge, this study is the first to apply both models to the TMJ simulation. The models incorporate information that cannot be captured by approaches based solely on landmarks \cite{KIM2013169}, selected regions of interest \cite{almashraqi2024recommendations}, or local distance metrics \cite{IKEDA2009}. Both simulation methods appear feasible; however, they yield slightly different results. The proposed methodology for quantifying and comparing these results revealed uncertainties in mandible positioning introduced by the semi-virtual approach and demonstrated how these uncertainties propagate into TMJ joint space simulations.

The iterative log-residual mean was used to obtain a representative rigid-body transformation without calculating the arithmetic mean of the individual matrix elements. Because the updates are performed using the group logarithm and exponential, each iteration remains a valid rigid-body transformation. The resulting mean describes the central transformation of the analyzed observations, while PCA was used to characterize the variability of their translational and rotational components. Concatenation of medical observations followed by PCA has also been used in biomechanical motion analysis \cite{federolf2016novel}, in this particular case with subject-specific centreing and normalization applied before pooling the data.

PCA revealed that the translational variability of the analyzed transformations was predominantly aligned with the latero-lateral direction. However, this finding should be interpreted with consideration of the structure of the analyzed dataset. The PCA included observations from all seven target mandible positions, and therefore the resulting covariance reflects both variability among repeated experimental realizations within individual positions and systematic differences between the target positions. These two sources of variability were not separated in the present analysis. Consequently, the observed latero-lateral predominance characterizes the combined dataset and cannot be attributed exclusively to an intrinsic directional characteristic of the modeling or repositioning procedure. Further investigations should distinguish within-position variability from between-position differences to clarify the extent to which the observed error patterns depend on the mandible position. Moreover, translational and rotational errors represent different physical quantities and are expressed in different units (mm and degrees, respectively). Their numerical magnitudes are therefore not directly comparable. Accordingly, the PCA results describe the directional characteristics of translational variability but do not establish the overall predominance of translational over rotational variability. A direct quantitative comparison between these components would require an appropriate common scale, for example, by expressing rotational deviations as equivalent linear displacements at a defined reference distance.

The observed pattern may be associated with several contributing factors, including residual bias in multimodal registration \cite{park2022creation,mangano2018combining}, systematic characteristics of occlusal splint geometry or positioning \cite{TOMAKA2025104527}, inaccuracies in defining the reference occlusal position \cite{kattadiyil2021what, IKEDA2009}, manufacturing and post-processing effects \cite{orgev2023effects}, and limitations of the measurement setup related to plaster models \cite{keating2008comparison} and rotary-table scanning \cite{man2019calibration}. Furthermore, translational and rotational errors represent different physical quantities expressed in different units (millimetres and degrees, respectively), and their numerical magnitudes are therefore not directly comparable. Accordingly, the present results do not establish the overall predominance of translational over rotational variability. A direct quantitative comparison would require an appropriate common scale, for example, by expressing rotational deviations as equivalent linear displacements at a defined reference distance.

The proposed methodology also depends on the certainty of all transformations linking the multimodal datasets \cite{TOMAKA2025110311,park2022creation}. Consequently, any error in determining the spatial relationship between the maxillary and mandibular teeth was propagated directly to the estimated positions of the articular surfaces within the TMJs. Residual errors in CBCT-surface-meshes registration \cite{park2022creation,park2020registration}, facial-motion registration \cite{TomakaG2019,TOMAKA2025110311}, or dental-model alignment \cite{keating2008comparison} can also affect the simulation of TMJ joint space changes. The accumulation of such errors may explain why the consistency error exhibited predominantly anisotropic variability in the translational component, whereas the model disagreements were not uniform across the target mandible positions. The regions exhibiting changes in joint space distances between the virtual and semi-virtual models were spatially non-uniform. This observation suggests that discrepancies between transformations may affect different regions of the joint to varying degrees and that these discrepancies can be characterized using combined rigid-body transformation and distance-based metrics since transformation-based analysis and surface-distance analysis provide complementary information \cite{li2025integrated,SHU2022107149}. The former characterizes the global rigid-body discrepancy between planned and achieved mandible positions \cite{li2025integrated}, while the latter reflects the local geometric agreement between surfaces \cite{SHU2022107149}. This dual representation is important because mandible positioning errors can affect the entire stomatognathic system and propagate to anatomically distant structures, including the mandibular condyles \cite{KIM2013169,li2025integrated}. Moreover, even small transformation discrepancies in mandible positioning can produce substantially larger changes in the spatial relationships between articular surfaces within the TMJs. However, the difference observed in this single-participant case cannot be generalized to broader clinical populations without further studies addressing the limitations of the present study.

One limitation of this study is that the virtual and semi-virtual models were compared with each other without an independent reference standard. Consequently, the findings reflect primarily internal consistency rather than providing an independent assessment of measurement accuracy. Therefore, further studies investigating the feasibility of applying this proof-of-concept methodology in clinical settings should be conducted using repeated CBCT imaging at different mandible positions \cite{henriques2012condylar,tsiklakis2004radiographic} repeated CBCT imaging with a physical phantom (e.g., an occlusal splint representing the mandible position) \cite{elbeialy2012interocclusal,nova2017positioning,vanderhelm2023validity}, or an alternative validated measurement method \cite{merema2022rotation,vanderhelm2023validity,xu2024dynamic}. 
Another limitation of this study relates to the fact that the anatomical structures were segmented only once, in a single participant, by a single experienced orthodontist. Although this approach is acceptable for a proof-of-concept study, repeated segmentation of multiple TMJs from multiple patients, together with independent segmentation by multiple observers, would be advisable before the proposed methodology is considered for clinical application. Such an approach would enable the assessment of intra-rater and inter-rater reliability \cite{mendez2019condylar,gorucu2019reliability}, segmentation repeatability \cite{verhelst2020validation}, and the potential impact of segmentation variability on joint space measurements \cite{holte2023holistic}. Furthermore, combining imaging data from multiple patients or biological models \cite{vanderhelm2023validity}, repeated imaging of the same patient/model in different mandible positions \cite{vanderhelm2023validity,henriques2012condylar}, repeated segmentation of anatomical structures by multiple observers \cite{mendez2019condylar,verhelst2020validation}, multiple splint fabrications \cite{reymus2020splints,cruzaraujo2025accuracy}, and repeated scanning of occlusal splint replicas placed on prepared models \cite{lee2024interocclusal,kakali2023bite} within a single experiment would enable the creation of a database necessary to quantify the contributions of individual intermediate experimental steps (including imaging, segmentation, occlusal splint fabrication, and occlusal splint scanning) to the uncertainties propagated to the TMJ-level outcomes. At this stage, the effects of occlusal splint fabrication, placement, and scanning can only be considered cumulatively, without the possibility of determining the individual contributions of occlusal splint geometry \cite{wang2022coverage}, material composition \cite{reymus2020splints}, material deformation \cite{maleki2024materials}, positioning errors \cite{schouman2015accuracy}, and scanning errors \cite{beidas2025printing} to the TMJ-level outcomes. However, in future validated cohort studies, the use of biological models rather than clinical patients should be considered, given the need for repeated imaging and the associated risks of cumulative exposure to X-ray radiation \cite{mallya2023recommendations}. Such biological models could include human cadavers \cite{coogan2018determination, coombs2019tmj}, animal cadavers \cite{lee2022animalmodels, jasinski2026accuracy}, or live animal models \cite{zhao2010rabbit, jasinski2024equine}, with the selection of an appropriate model guided by relevant ethical considerations.

Following further research addressing the limitations identified in the present study, the proposed methodology may prove potential clinical utility in evaluating patient-specific geometrically related positions of the mandible and TMJs articular surfaces \cite{li2025integrated}. In future validated cohort studies, it should also be considered that, in the proposed approach, the joint space distance was calculated as the shortest distance between the surface meshes of the mandibular condyle and the glenoid fossa. This approach differs from conventional measurements of TMJ joint space performed on CBCT images. In clinical studies, bilateral measurements of the anterior joint space (AJS) \cite{bk2025quantitative} or anterior space distance (AS) \cite{schnabl2018cbct}, superior joint space (SJS) \cite{bk2025quantitative} or superior space distance (SS) \cite{schnabl2018cbct}, and posterior joint space (PJS) \cite{bk2025quantitative} or posterior space distance (PS) \cite{schnabl2018cbct} are typically performed manually using the true sagittal view of each TMJ \cite{bk2025quantitative, schnabl2018cbct}. These measurements are based on predefined anatomical landmarks, such as the superior aspect of the glenoid fossa and the most prominent points on the anterior and posterior aspects of the mandibular condyle \cite{bk2025quantitative, schnabl2018cbct}, between which auxiliary measurement lines are drawn \cite{bk2025quantitative}. In studies employing machine learning-based approaches, bilateral measurements of the 3D superior joint space are performed by calculating the spatial distance between the apex of the mandibular condyle and the corresponding articular surface of the glenoid fossa \cite{mackie2022quantitative, al2024comprehensive, al2025interpretable}. The approach adopted in the present study is conceptually similar to the 3D measurement approaches used in these studies \cite{mackie2022quantitative, al2024comprehensive, al2025interpretable}; however, because it calculates the shortest distance between the meshes representing the opposing articular surfaces, it is more closely related to the distances measured in anterior location (AJS/AS) than in superior (SJS/SS) or posterior (PJS/PS) \cite{bk2025quantitative, schnabl2018cbct}. Taking these methodological differences into account, the proposed approach may be particularly useful when repeated radiographic imaging of the TMJ in different mandible positions is undesirable due to the associated exposure to X-ray radiation \cite{mallya2023recommendations}. In healthy adults images with teeth in the neutral position, the mean AJS/AS distance has been reported to range from 2.29 mm (AJS) \cite{bk2025quantitative} to 2.58 mm (AS) \cite{schnabl2018cbct}. In the present study, the joint space distance in the neutral position was 2.37 mm for the left TMJ and 2.56 mm for the right TMJ, both within the reported normal range. No analogous CBCT-based measurements performed in the open-mouth position have been reported in the available literature. Case reports describing the use of occlusal splints for occlusal treatment have reported measurements of the AJS \cite{jiang2026axiography} / AS \cite{nota2021workflow}, SJS \cite{jiang2026axiography} / SS \cite{nota2021workflow}, and PJS \cite{jiang2026axiography} / PS \cite{nota2021workflow}. However, although pre- and post-treatment imaging was performed, the TMJs were always imaged with the mandible in a neutral position. In a cohort study involving mandibular motion modeling, it was demonstrated that, during mouth opening, the mandibular condyle covers a pathway ranging from $2.90 \pm 2.59$ mm to $7.23 \pm 3.16$ mm (mean $\pm$ SD), with the highest value occurring during maximum mouth opening. However, the maximum mouth-opening range was not reported and CBCT-based measurements of joint-space distances were not performed with the mouth open \cite{YANG2013}. It can therefore be conservatively estimated that the joint space distance values reported in the present study are close to the upper range of values associated with mandibular motion. Nevertheless, since these results of mathematical models do not account for the restrictions imposed on TMJ movement by the joint capsule and ligaments and have not been validated against multiple CBCT imaging, their anatomical relevance should be interpreted with caution.
Interestingly, changes in joint space distance have been reported in temporomandibular disorders (TMDs), e.g., both TMJ osteoarthritis \cite{mackie2022quantitative, al2024comprehensive, al2025interpretable} and bruxism \cite{bk2025quantitative}, making it an important radiographic indicator of these conditions \cite{schiffman2014diagnostica}. Therefore, the observed 1.46 mm difference in joint space distance between the models used for TMJ joint space simulations falls within a measurable range and may be clinically meaningful, as it could influence the assessment of joint space narrowing or widening and, consequently, final diagnosis. However, the generalizability of the present single-participant study is limited, as it serves as a methodological proof-of-concept that demonstrates how mandible positions can be propagated to a simulated TMJ model and how model-based uncertainty can be quantified. Further extensive research involving larger and more diverse participant cohorts is required before the methodology can be applied to the patient-specific diagnosis of TMDs \cite{schiffman2014diagnostica} or follow-up assessment of treatment outcomes \cite{keller2023patientreported}.

\section{Conclusions}

This study presented the feasibility of the transformation tree-based methodology for the propagation of mandible positions to simulated TMJ models and the quantification of the model-based uncertainty. In the case of studied single participant, both developed models - the virtual and the semi-virtual - simulated bilateral symmetrical changes in joint space distance in mandible positions involving only the mouth opening, and bilateral asymmetrical changes in positions involving the mouth opening with lateral deviation. Thus, changes in mandible positions can be propagated to TMJ articular surface positions using rigid-body transformations, although the results are influenced by the simulation model applied. Quantification of the model-dependent effect revealed the variability of the consistency error which dominated in the translation component over the rotation component of the rigid-body transformation, and showed anisotropy in the latero-lateral direction within common spatial coordinate system. The model disagreements were not uniform across the target mandible positions, indicating that transformation discrepancies can affect different regions of the joint to varying degrees. 
In this single-participant study, small transformation discrepancies in mandible position led to significantly higher changes in the spatial relationships of the articular surface within the TMJ. Although our study showed that these changes can be characterized using a combination of rigid-body transformation and distance-based metrics, the numerical findings should not be generalized to broader clinical populations given the proof-of-concept nature of this study.

\backmatter

\bmhead{Funding}
The publication was (co)financed by Science development fund of the Warsaw University of Life Sciences—SGGW.

\bmhead{Data availability}
The original multimodal dataset, including medical images and 
anatomical surface meshes, is not publicly available due to 
privacy and ethical considerations. The derived numerical data 
required to reproduce the results presented in the tables are 
publicly available as part of the reproducibility package 
deposited in Zenodo \cite{scripts_zenodo}.

\bmhead{Code availability}
Parts of the software infrastructure used in this study are 
publicly available through the dpVision platform 
\cite{dpvision_soft} and the associated SplintMaker plugin 
\cite{splintmaker_soft}. The analysis scripts required to 
reproduce the numerical results presented in the tables are 
included in the reproducibility package deposited in Zenodo 
\cite{scripts_zenodo}. Additional technical information may 
be obtained from the corresponding author upon reasonable request.

\bmhead{Author contributions}
\textbf{A.A.T.} is responsible for the conceptualization of the study, methodology development, multimodal data acquisition and integration, and preparation of the initial manuscript draft.
\textbf{K.D.} is responsible for the formal analysis, statistical methodology, mathematical interpretation of the results, and project supervision.
\textbf{M.T.} is responsible for the clinical and medical aspects of the study, supervision of the medical data acquisition process, interpretation of the biomedical context, and critical review of the manuscript.
\textbf{E.K.} is responsible for interpretation of the medical context, and critical review of the manuscript.
\textbf{M.D.} is responsible for the formal analysis, critical reviewing of the initial version of the manuscript, and preparing the final version of the manuscript.
\textbf{D.P.} is responsible for software development, implementation of visualisation and computational methods, data processing, participation in the statistical analysis, and manuscript writing, editing, and review.

\section*{Declarations}

\bmhead{Conflicts of interest}
The authors declare that they have no conflict of interest.

\bmhead{Ethics declarations}
This study was performed in accordance with the principles of the Declaration of Helsinki. Approval was granted by the Bioethics Committee of the Silesian Medical Chamber in Katowice, Poland (Resolution No. ~33/2021, 22 September 2021). All the procedures were performed in accordance with relevant guidelines and regulations.

\bmhead{Consent to participate}
All participants provided written informed consent to participate in the study. Specifically, the archival multimodal data and a dental plaster model were obtained from a single adult participant, who provided written informed consent for their use. 

\bmhead{Consent for publication}
The participant consented to the use of the archival multimodal data and dental plaster model for scientific publication. All authors consented to publication of the manuscript.

\bibliography{newrefs.bib}

\end{document}